\documentclass[draftcls]{IEEEtran}
\usepackage{cite}      % Written by Donald Arseneau
\usepackage{graphicx}  % Written by David Carlisle and Sebastian Rahtz

\usepackage{subfigure} % Written by Steven Douglas Cochran

\usepackage[nomarkers]{endfloat}
\let\MYoriglatexcaption\caption
\renewcommand{\caption}[2][\relax]{\MYoriglatexcaption[#2]{#2}}
\begin{document}
\title{Ground tracking for improved landmine detection in a GPR system}
\author{Li~Tang,
        Peter~A.~Torrione,~\IEEEmembership{Member,~IEEE,}
        Cihat~Eldeniz,
        and~Leslie~M.~Collins,~\IEEEmembership{Senior~Member,~IEEE}% <-this % stops a space
%\thanks{Manuscript received January 20, 2002; revised November 18, 2002.
%        This work was supported by the IEEE.}% <-this % stops a space
\thanks{The authors are with the Dept. of Electrical and Computer Engineering at Duke University, Durham, NC 27708 USA.}}
% note the % following the last \IEEEmembership and also the first \thanks - 
% instance, "A\textbf{} \textbf{}B" will typeset as "A B" not "AB". If you want
% "AB" then you have to do: "A\textbf{}\textbf{}B"
\markboth{Submitted to IEEE Transactions on Geoscience and Remote Sensing}{Shell \MakeLowercase{\textit{et al.}}: IEEEtran.cls for Journals}
% *** Note that you probably will NOT want to include the author's name in ***
% *** the headers of peer review papers.                                   ***

% If you want to put a publisher's ID mark on the page
% (can leave text blank if you just want to see how the
% text height on the first page will be reduced by IEEE)
%\pubid{0000--0000/00\$00.00~\copyright~2002 IEEE}

%\specialpapernotice{(Invited Paper)}

\maketitle

\begin{abstract}
Ground penetrating radar (GPR) provides a promising technology for accurate subsurface object detection. In particular, it has shown promise for detecting landmines with low metal content. However, the ground bounce (GB) that is present in GPR data, which is caused by the dielectric discontinuity between soil and air, is a major source of interference and degrades landmine detection performance. To mitigate this interference, GB tracking algorithms formulated using both a Kalman filter (KF) and a particle filter (PF) framework are proposed. In particular, the location of the GB in the radar signal is modeled as the hidden state in a stochastic system for the PF approach. The observations are the 2D radar images, which arrive scan by scan along the down-track direction. An initial training stage sets parameters automatically to accommodate different ground and weather conditions. The features associated with the GB description are updated adaptively with the arrival of new data. The prior distribution for a given location is predicted by propagating information from two adjacent channels/scans, which ensures that the overall GB surface remains smooth. The proposed algorithms are verified in experiments utilizing real data, and their performances are compared with other GB tracking approaches. We demonstrate that improved GB tracking contributes to improved performance for the landmine detection problem.
\end{abstract}


% Will be ignored for other modes.
\IEEEpeerreviewmaketitle

\section{Introduction}
% \PARstart{A}{demo} file is ....
% \PARstart{A}{}demo file is ....
% \PARstart{T}{his demo} file is ....
% 
\PARstart{G}{round} penetrating radar (GPR) is a promising sensor modality for the landmine detection problem, particularly since these sensors are capable of detecting landmines with low metal content. Much research has recently focused on landmine detection with GPR systems (e.g. \cite{Gader:HMM, Peter:feature, Ho:Handheld, Potin:SVM, Zhu:Hyperbolas, Merwe:clutterReduction, Xu:arrayGPR, Delbo:fuzzy, Ho:frequency, Carevic:Wavelet}). In landmine detection applications, the goal is to flag the positions of all landmines with a minimum number of false alarms. This means that features that can distinguish landmines from background clutters have to be formulated and extracted. Historically, a combination of features from both the time-domain and the frequency-domain have been used to achieve low false alarm rates. However, in spite of the promising results that have been obtained using GPR sensors, very challenging problems remain unsolved. Difficulties remain that are associated with many factors, such as different landmine types, soil conditions, temperature and weather. Moreover, it is difficult to find a consistent and robust model to describe the features associated with landmine signatures under all combinations of the factors described above.

One issue that has been a problem for almost all landmine detection algorithms is eliminating the radar return from the ground, or the ``ground bounce" (GB), as it is a significant source of false alarms. For example, Fig.~\ref{fig:example} shows two GPR images representing vertical slices of the ground collected from government-sponsored test sites. The x-axis represents the down-track direction while the y-axis shows the time/depth dimension. The bright curves that track along the down-track direction are the GB responses. There are 6 landmines appearing as sets of hyperbolas in Fig.~\ref{fig:example}(a). Landmine detection algorithms are designed to find areas of interest in the radar data caused by the dielectric discontinuities between soil and landmines. As the GB is produced by the dielectric discontinuities between the ground and the air, it is a significant source of background clutter. It is therefore generally accepted that the GB must be detected and removed. Inaccurate location of the GB can also impact feature extraction. Fig.~\ref{fig:example}(b) shows data collected from the same radar in cold weather with snow covering the ground. Clearly, the true GB is more difficult to discern in Fig.~\ref{fig:example}(b) than in Fig.~\ref{fig:example}(a).

A number of algorithms have been proposed for GB tracking and clutter removal in order to increase the accuracy of landmine detection. Clutter due to the GB is modeled using a parametric model in \cite{Merwe:clutterReduction}. In \cite{Delbo:fuzzy}, a wavelet denoising approach is proposed to decrease the importance of homogeneous parts of the data (the background) against those where high frequencies are dominant (the edges). A signal model \cite{Brunzell:shallowly} exploits the different properties of the backscattered signals from target and ground surface. Wu et al. estimate and eliminate the GB from a given scan using a shifted and scaled version of an adaptively estimated reference GB \cite{Wu:AdaptiveGB}. The time shift and amplitude scaling of the ground surface clutter are identified by correlation with the ideal flat ground for clutter removal in \cite{Rappaport:roughGround}. An adaptive filter in \cite{Kempen:parameters} estimates the correlated clutter parameters. These approaches performed well in relatively benign conditions, but may encounter difficulties in more difficult scenarios such as those depicted in Fig.~\ref{fig:example}(b).

The main challenge for GB tracking in the real world is that there are a variety of ground conditions, such as soil, sandy, gravel, asphalt surfaces, or ground covered with vegetation. These various ground surfaces often result in significant anomalies unrelated to the presence of a landmine. These anomalies are inhomogeneous and the statistical properties of the GB responses may vary with position \cite{Merwe:clutterReduction}. GB response characteristics are also influenced by weather conditions, such as soil humidity, rain and snow. Another important factor that must be considered in GB tracking is the latency issue. In this study, the Wichmann/Niitek GPR system was used to collect data \cite{Peter:feature}. Unlike hand-held GPR systems \cite{Ho:Handheld}, the radar sensor is mounted some distance in front of a vehicle. The detection system should be able to produce reliable alarms before the vehicle wheel base encounters the landmines in real applications. The number of data scans available for processing is thus limited by the distance between the sensor and the vehicle. In essence, an ideal GB tracking system should be able to perform robustly in all kinds of road and weather conditions with minimum latency.

In this paper, we provide a comparison of four different GB tracking approaches. First, we discuss three computationally simple GB trackers, which are referred to as the ``global maximum", ``constrained maximum" and ``Kalman filter" (KF) respectively. Then we formulate the GB tracking problem using a particle filter (PF) framework, which is a technique used to estimate the state of nonlinear / non-Gaussian stochastic systems \cite{Gordon:Tutorial}. The PF allows more flexibility than a KF, which requires that both the system and observation models are linear functions with additive Gaussian process noise and measurement noise, respectively \cite{Gordon:Tutorial}. For the PF, in contrast, no such constraints exist. Therefore, the PF is potentially a better choice for GB tracking problems, since it is hard to relate the GB locations to the observations with a linear function due to the inhomogeneous GB signatures \cite{Rappaport:roughGround}. The pros and cons of these GB trackers are analyzed accordingly with experimental results.

The remainder of this paper is organized as follows. The GB tracking problem is formulated using three computationally simple GB trackers in Section~\ref{Sec:formulation}. In Section~\ref{Sec:GBtrackingPF}, after a brief review of the particle filter, the GB tracking problem is placed within this framework and detailed implementation is presented with improvements on generic PF schemes described. Section~\ref{Sec:comTable} compares and summarizes both the KF and the PF formulations in a table. The proposed algorithms are verified and compared with those simple GB trackers in Section~\ref{Sec:experiments} utilizing real data collected at government-sponsored test sites. Finally, some concluding comments are given in Section~\ref{Sec:conclusion}.

\section{Problem formulation}
\label{Sec:formulation}

In this section, the GB tracking problem is introduced with three computationally simple trackers proposed.

\subsection{Problem statement and computationally simple GB trackers}
\label{Sec:simpleTrackers}

The Wichmann/Niitek GPR system (frequency band: 200MHz-7GHz) used in this study has the GPR antenna mounted in front of a vehicle. The antenna has 24 transmit/receiver pairs (channels) spanning 1.2 meters cross-track. As the vehicle moves along the test lanes (down-track direction), a 3D volume of data is produced scan by scan with dimensions time / frequency / depth, cross-track (XT) and down-track (DT). In the DT direction, all channels are sampled once every 5cm and at each location, a 415-element time/depth domain vector is recorded. Therefore, the radar data is represented as a 3D matrix with dimensions: ${N}_d\times{N}_{CH}\times{N}_{DT}$, where $N_d=415$ is the number of depth, or time, samples, $N_{CH}=24$ is the number of channels and $N_{DT}$ is the number of the DT scans in a particular run. At one DT/XT location, the radar response in time/depth is called an A scan with dimensions ${N}_d\times{1}$, and all of the A scans in one channel along the DT direction comprise a B scan with dimensions $N_d\times{N}_{DT}$, which represents the radar responses of a vertical slice of the ground.

\begin{figure}%performanceComparison.m
	\centering
	$\begin{array}{c}
		\includegraphics[height=1.6in]{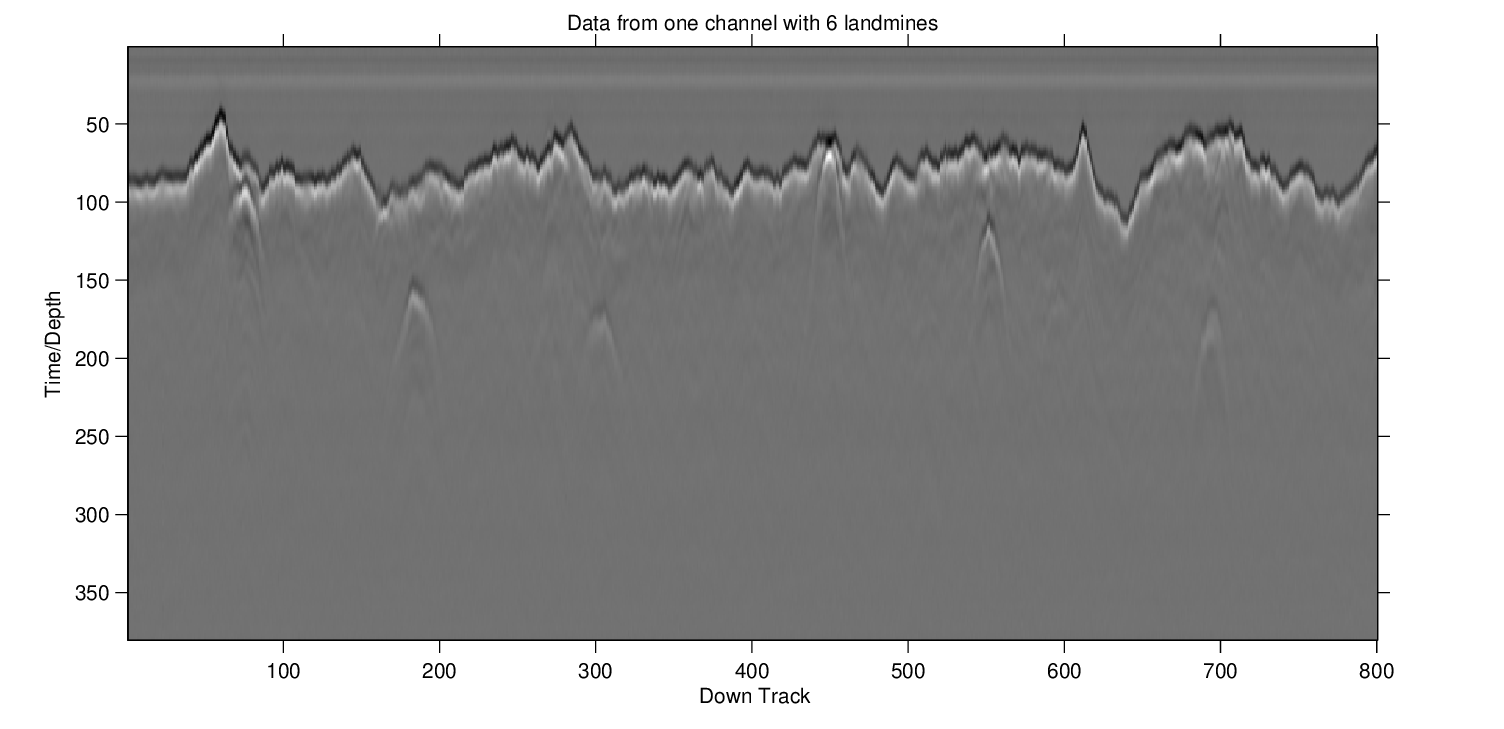}\\%belvoir_GM.eps,CHnum=6
		(a)\\
		\includegraphics[height=2in]{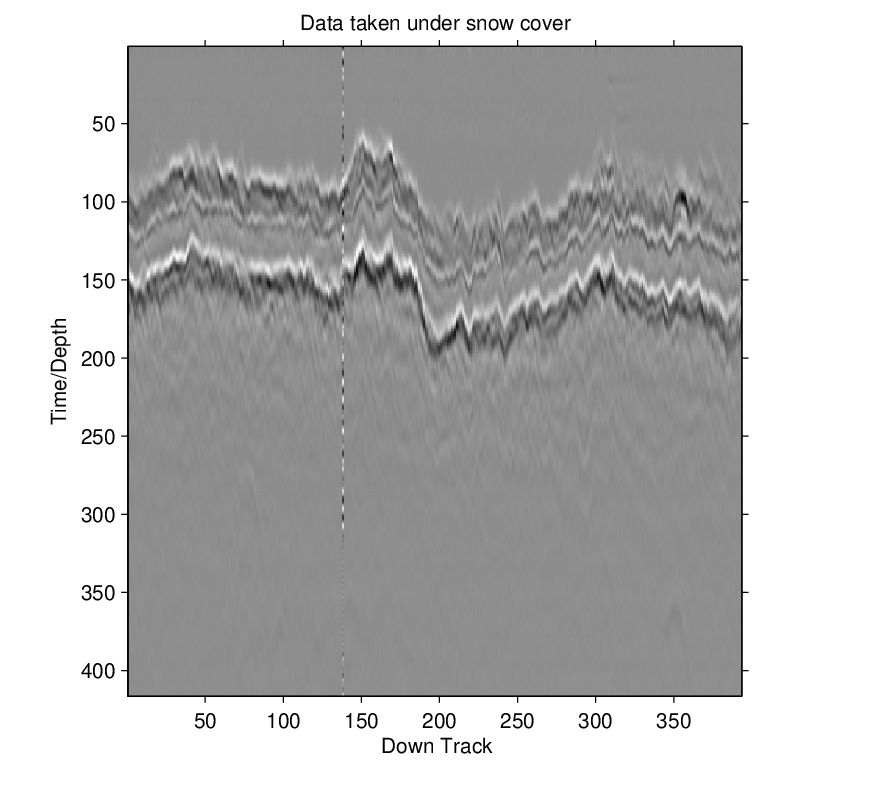}\\%snow_GM.eps,CHnum=3
		(b)
	\end{array}$
	\caption{Two typical examples of the GPR data: (a) One channel with 6 landmines located at down-track scans of 76, 185, 306, 450, 553 and 695; (b) One channel with data taken under snow cover. GPS interference is present near scan 150.}
	\label{fig:example}
\end{figure}

Due to the dielectric discontinuities between the ground and the air, there is a large GPR response in each A scan from the ground/air interface. As a simple GB tracker, GB locations can be roughly estimated by finding the maximum response along each of these A scans, which is denoted as $GB_{max}$. This approach will be referred to as the ``global maximum" method. In most cases, particularly in benign environments, the ground/air interface does in fact generate the maximum response in the GPR signal. However, there are a number of cases where the maximum response is generated by other factors, such as the interface between snow and the air (Fig.~\ref{fig:example}(b)) or surface metallic objects and the air. Other subsurface anomalies can also be problematic for a simple GB tracker. For example, in Fig.~\ref{fig:example}(b) near scan 150, data indicative of GPS (Global Positioning System) interference can be observed. These anomalies cause GB tracking based on a global maximum to ``jump" from one location to another, which significantly impacts the accuracy of landmine detection, particularly if the interferences occur in the vicinity of a landmine.

%\WichmannData\belvoir_metal_data\Metal_Mine_IED_0-3-6_Run1_155CT.mdr
%\WichmannData\CRREL_Data_From_Pete\2005_03_01_data\torrione_data\CARCASS\...
%Lane1_M1_15cm_g1_0_20_Snow_no_water.mdr

An alternative yet still computationally simple GB tracker is based on finding local maximum responses, which will be referred to as the ``constrained maximum" method. For every A scan, this algorithm searches for the maximum radar response in a ``safe" neighborhood centered the GB location of the previous estimate, where the size of the neighborhood $W$ is defined to be proportional to the standard deviation of the GB locations estimated so far:
\[
W_{dt}=min[W_{max},\mbox{ }\alpha\hspace{-0.02in}\cdot\hspace{-0.02in}{STD}(GB_1,\cdots,GB_{dt})]
\]

\noindent where $STD(\cdot)$ represents the standard deviation of a vector. The size of the search window is bounded by $W_{max}$ and $\alpha$ is a pre-defined constant. For a given data set, if the size of the search window $W_{dt}$ is set properly, both the accuracy and the computational efficiency of the GB tracking can be optimized. However, if the computed standard deviation of a specific data set is small, the GB location is dependent heavily on its previous estimate. This may cause difficulties for the tracker to recover from wrong GB locations. On the other hand, if there are some noises existing in the previous estimates, $W_{dt}$ is prone to getting larger and larger until its upper bound $W_{max}$ is reached, which is same as tracking with a constant search window. Moreover, $\alpha$ is a parameter to be chosen properly. So if a variety of data sets are considered together, it is difficult to make the size of the search window fit all circumstances, and this lack of robustness degrades the overall performance.

\subsection{Kalman filtering formulation}

A Kalman filter \cite{Welch:Kalman} potentially provides a more accurate GB tracker than the approaches considered in Section~\ref{Sec:simpleTrackers}. A KF formulation was implemented with constraints enforced by the linear system/observation models:
\begin{eqnarray}
\label{Eq:Kalman1}
\textbf{x}_k=f_k(\textbf{x}_{k-1},\textbf{v}_k)=\textbf{F}\textbf{x}_{k-1}+\textbf{v}_k\\%\textbf{B}_k\textbf{u}_k+
\label{Eq:Kalman2}
\textbf{z}_k=h_k(\textbf{x}_k,\textbf{n}_k)=\textbf{H}\textbf{x}_k+\textbf{n}_k%\\%\nonumber
\end{eqnarray}

\noindent The state vector is supposed to be able to fully summarize the past of the dynamic system. Hence at every DT/XT location it is constructed as:

\[
\textbf{x}_k=[GB_{ch}^k\mbox{ }GB_{ch-1}^k\mbox{ }GB_{ch+1}^k\mbox{ }{GB'}_{ch}^k\mbox{ }{GB'}_{ch-1}^k\mbox{ }{GB'}_{ch+1}^k]^T
\]

\noindent with $GB_{ch}^k$ representing the GB location at channel $ch$ and scan $k$. The state vector consists of two parts. The first three items are the GB locations of three adjacent channels ($ch$, $ch-1$ and $ch+1$) at scan $k$ while the second three items are their discrete derivatives, i.e. ${GB'}_{ch}^k=GB_{ch}^k-GB_{ch}^{k-1}$.

The state transition matrix $\textbf{F}$ in the system model Eq.~\ref{Eq:Kalman1} is chosen in such a way that prediction of the current GB location $GB_{ch}^k$ is based on the mean of three adjacent locations at its previous scan $GB_{ch-1}^{k-1}$, $GB_{ch}^{k-1}$ and $GB_{ch+1}^{k-1}$. Its derivative ${GB'}_{ch}^k$ depends on the mean of ${GB'}_{ch-1}^{k-1}$, ${GB'}_{ch}^{k-1}$ and ${GB'}_{ch+1}^{k-1}$. So we have
\begin{eqnarray}
\textbf{F} & = &
\left[ \begin{array}{cccccc}
1/3 & 1/3 & 1/3 & 1/3 & 1/3 & 1/3\\
1/3 & 1/3 & 1/3 & 1/3 & 1/3 & 1/3\\
1/3 & 1/3 & 1/3 & 1/3 & 1/3 & 1/3\\
0 & 0 & 0 & 1 & 0 & 0\\
0 & 0 & 0 & 0 & 1 & 0\\
0 & 0 & 0 & 0 & 0 & 1
\end{array} \right]
\end{eqnarray}

As the limitation of its linearity nature, it is hard to incorporate the original GPR signature in the observation model. So GB estimates from the global maximum $GB_{max}^k$ at three adjacent channels ($ch$, $ch-1$ and $ch+1$) are taken to compose the measurement $\textbf{z}_k$ with
\begin{eqnarray}
\textbf{H} & = &
\left[ \begin{array}{cccccc}
1 & 0 & 0 & 0 & 0 & 0\\
0 & 1 & 0 & 0 & 0 & 0\\
0 & 0 & 1 & 0 & 0 & 0
\end{array} \right]
\end{eqnarray}

Both the process noise ($\textbf{v}_k\sim{N}(0,\textbf{Q}_k)$) and the observation noise ($\textbf{n}_k\sim{N}(0,\textbf{R}_k)$) obey zero mean Gaussian distributions, where $\textbf{Q}_k$ and $\textbf{R}_k$ are the corresponding covariance matrices. The process noise covariance $\textbf{Q}_k$ is chosen to be low and constant, which means given the right GB locations and their derivatives of three neighboring channels, we are pretty sure there are not going to be jumps. This is equivalent to assuming the ground is fairly smooth. If there is an abrupt jump on the ground, it may take some time for the KF to find the right track due to the low process noise. But as long as the results from the global maximum $GB_{max}$ stay at the right locations steadily, the KF will eventually recover from those errors. In contrast, the observation noise covariance $\textbf{R}_k$ is supposed to be high and variable, because we are not sure if the result from the global maximum is the real GB location. But our confidence increases if GB variations across channels give similar values. Therefore, the observation noise covariance $\textbf{R}_k$ is set to be adaptive with the standard deviation of GB variations across channels followed by a smoothing filter.

Experimentally, one would expect the KF to perform better than the global maximum since the ``history" of the state is considered during the GB estimation process. Specifically, the abrupt ``jumps" due to random effects are generally removed. However, the KF approach cannot improve the performance when the results from the global maximum $GB_{max}$ remain in the wrong place for a number of consecutive scans, as this approach adopts $GB_{max}$ as the observation without taking the original 3D radar data into consideration at the filtering stage. That is to say, the KF formulation is the minimum mean squared error (MMSE) estimate to $GB_{max}$ with a Gaussian observation noise \cite{Welch:Kalman}. Moreover, the KF predicts and updates the prior GB estimates using linear models, which may not accommodate different ground and weather conditions.

%%%%%%%%%%%%%%%%%%%%
%For the entire data set, there are $N_{DT}\times{N}_{CH}$ locations to be processed, which is referred to as ``time step" $k$ in our state-space estimation system.
%All of the above mentioned drawbacks of these existing approaches motivate us to introduce particle filter into the GB tracking problem.
%%%%%%%%%%%%%%%%%%%%

\section{GB tracking using particle filter}
\label{Sec:GBtrackingPF}

In this section, a brief review of the generic PF algorithm is provided. To build an efficient PF framework for the GB tracking problem, two key issues have to be addressed properly. One issue is related to drawing good prior samples so that they are placed at the important regions of the posterior, which relies on a state prediction according to the system dynamics model. The other issue involves accurately evaluating the importance of these samples upon the receipt of the new observation, and updating the particles to obtain the posterior estimate. This involves a likelihood computation derived from the observation model.

\subsection{PF basics}

Suppose the evolution of a hidden state $\{\textbf{x}_k, k\in{\textbf{N}}\}$ to be estimated in space is a first order Markov system given by:
\begin{equation}
\label{Eq:system}
\textbf{x}_k = f_k(\textbf{x}_{k-1},\textbf{v}_k)
\end{equation}

\noindent where $f_k$ describes the state transition from time step $k-1$ to $k$ and $\{\textbf{v}_k, k\in{\textbf{N}}\}$ is i.i.d. process noise. The hidden state $\textbf{x}_k$ is estimated from the observations $\textbf{z}_k$ defined by:
\begin{equation}
\label{Eq:measurement}
\textbf{z}_k = h_k(\textbf{x}_k,\textbf{n}_k)
\end{equation}

\noindent where $h_k$ is the observation model relating the state $\textbf{x}_k$ to the measurement $\textbf{z}_k$. These two models have similar forms as those mentioned in the KF framework (Eq.~\ref{Eq:Kalman1} and Eq.~\ref{Eq:Kalman2}). However, both $f_k(.)$ and $h_k(.)$ can be any functions in the PF formulation instead of the linear constraints enforced by a Kalman filter. In a PF formulation, at each time step $k$, the posterior distribution of the state $\textbf{x}_k$ can be approximated by a set of particles $\{\textbf{x}_k^i, i=1,\ldots,N_p\}$ with associated weights $\{w_k^i, i=1,\ldots,N_p\}$:
\begin{equation}
\label{Eq:particle}
p(\textbf{x}_k|\textbf{z}_{1:k}) = \sum_{i=1}^{N_p}{w}_k^i\delta(\textbf{x}_k-\textbf{x}_k^i)
\end{equation}

\noindent where $\delta(.)$ is the Dirac delta function. $N_p$ is the number of the particles and the weights $w_k^i$ are normalized such that $\sum_i{w}_k^i=1$. Each particle is a sample with $\textbf{x}_k^i$ representing one possible state. The associated weights represent the ``importance" or a kind of ``support" those particles had from the current observation $\textbf{z}_k$. In estimating the state at a given time step, those particles with large weights have larger contributions to the state estimate than those with smaller weights. The state estimate $\hat{\textbf{x}}_k$ is either approximated by the minimum mean square error (MMSE) estimate or by the maximum a posteriori (MAP) estimate \cite{Zhou:Adaptive}:
\begin{equation}
\label{Eq:MMSE_MAP}
\hat{\textbf{x}}_k=\sum_{i=1}^{N_p} w_k^i\textbf{x}_k^i\mbox{, or }arg\mbox{}\max_{\textbf{x}_k}w_k^i.
\end{equation}

\noindent Since the posterior distribution is not represented by a closed-form function, it can approximate any distributions if given sufficient particles.

The weights are assigned to each particle using the principle of importance sampling \cite{Gordon:Tutorial}. Firstly, samples (particles) are drawn from a proposal distribution $q(x)$, which is usually approximated by the prior importance function for simplicity:
\begin{equation}
\label{Eq:proposal}	
\bar{\textbf{x}}_k\sim{q}(\textbf{x}_k|\textbf{x}_{1:k-1},\textbf{z}_{1:k})\approx{p}(\textbf{x}_k|\textbf{x}_{k-1})
\end{equation}

\noindent where $q(x)$ is also known as the importance function. To distinguish from the true state $\textbf{x}_k^*$ and the posterior estimate $\hat{\textbf{x}}_k$, $\bar{\textbf{x}}_k$ is used to denote the prior estimate. Eq.~\ref{Eq:proposal} is a one-step-ahead prediction of the hidden state $\textbf{x}_k$ \cite{Andrieu:invited}. With the new observation $\textbf{z}_k$ available, the recursive weight computation can be simplified by applying the Bayesian formula as follows \cite{Rui:UnscentedPF}:
\begin{equation}
\label{Eq:weight}	
w_k^i=w_{k-1}^i\frac{p(\textbf{z}_k|{\textbf{x}}_k^i)p({\textbf{x}}_k^i|\textbf{x}_{k-1}^i)}{q({\textbf{x}}_k^i|\textbf{x}_{1:k-1}^i,\textbf{z}_{1:k})}\approx{w}_{k-1}^i{p}(\textbf{z}_k|{\textbf{x}}_k^i)
\end{equation}

\noindent where the conditional PDF of $\textbf{z}_k$ given $\textbf{x}_k$, $p(\textbf{z}_k|{\textbf{x}}_k)$ is the likelihood function.

Basically, the recursive estimation process consists of two stages: \emph{predict} and \emph{update}. Give a set of particles with associated weights $\{\textbf{x}_{k-1}^i,w_{k-1}^i\}_{i=1}^{N_p}$ at time step $k-1$, a prior distribution of the state at time step $k$ is predicted using the state-space transition model (Eq.~\ref{Eq:system}), which essentially means to apply the system dynamics $f_k$ to $\{\textbf{x}_{k-1}^i\}$ with appropriate process noise $\textbf{v}_k$ to obtain $\bar{\textbf{x}}_k$. When a new observation $\textbf{z}_k$ is available, the weights at the previous time step $\{w_{k-1}^i\}$ are updated according to Eq.~\ref{Eq:weight} using the observation model (Eq.~\ref{Eq:observation}). The posterior distribution of the estimated state at current time step $k$ is therefore approximated by the newly produced particle set with their weights $\{\textbf{x}_k^i,w_k^i\}_{i=1}^{N_p}$ (Eq.~\ref{Eq:particle} and Eq.~\ref{Eq:MMSE_MAP}). To summarize, particles are propagated through the system evolution model in the predict stage, which includes a deterministic component of the system dynamics and a random component of the process noise \cite{Blake:CONDENSATION}. With the new observation from the current time step, the prediction is refined in the update stage to provide the posterior estimate.

One problem with the PF is that, usually after several iterations, only few particles have significant weights while the weights of the other particles become negligibly small. This means most of the computations are devoted to particles whose contributions to the overall state estimation are very small. This problem is referred to as the \emph{degeneracy phenomenon} and is usually avoided by \emph{resampling} \cite{Gordon:Tutorial}, which generates a new particle set from the current one with probabilities proportional to their weights. After resampling, all of the new particles are assigned with equal weights, which means they are properly placed over the important regions of the posterior \cite{Wang:Modular}. This provides a good starting point for the state prediction at the next time step.

\subsection{Particle filtering formulation of GB tracking}

Firstly, we analyze the nature of the GB tracking problem and formulate a solution using the PF framework. Both the system model and the observation model are established accordingly.

Suppose the evolution of the GB locations at each DT/XT position is the hidden state $\{\textbf{x}_k, k\in{\textbf{N}}\}$ to be estimated in Eq.~\ref{Eq:system}, where $f_k$ describes the variation of GB locations from scan $k-1$ to $k$. The GB tracking problem is thus formulated as estimating $\textbf{x}_k$ from Eq.~\ref{Eq:measurement}, where $h_k$ is the observation model relating the GB location $\textbf{x}_k$ to the measured radar signal $\textbf{z}_k$. For each time step $k$, $\textbf{z}_k$ is a ${N}_d\times{1}$ vector in the ${N}_d\times{N}_{CH}\times{N}_{DT}$ matrix. The i.i.d. measurement noise $\{\textbf{n}_k, k\in{\textbf{N}}\}$ is assumed to obey a zero mean Gaussian distribution $N(0,\sigma_n^2)$. The GB location at every DT/XT position is taken as the one-dimensional state vector $\textbf{x}_k$ in this formulation, since the computational intensity of the PF increases exponentially with the increase in the dimension of the state vector.

\begin{figure}
	\centering
	\includegraphics[height=1.65in,width=3in]{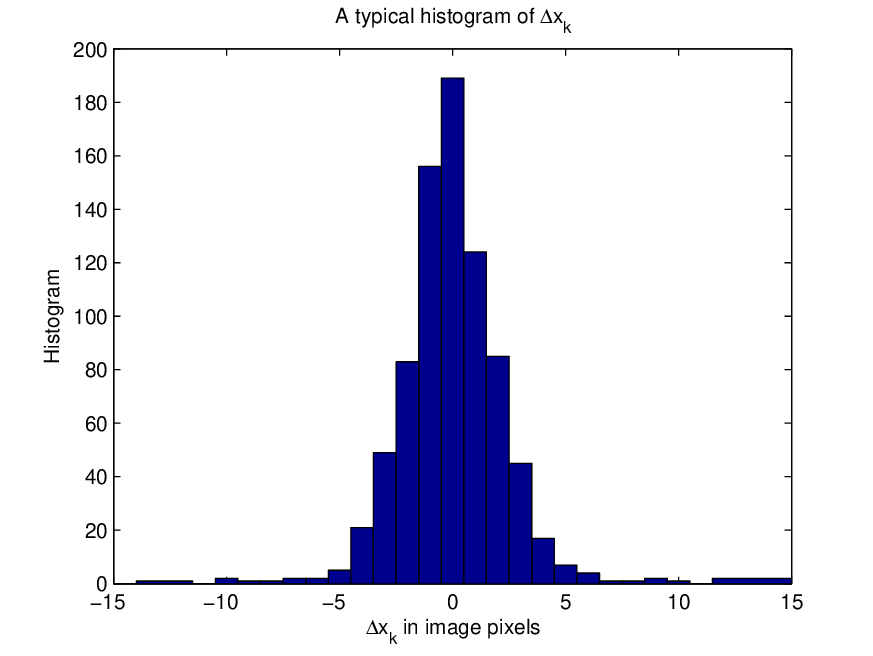}%myAligndata3D.m,belvoir,CHnum=4
	\caption{A typical distribution of $\Delta\textbf{x}_k$ in the experimental data set.}
	\label{fig:GBvariation}
\end{figure}

To obtain an appropriate state transition model, we analyze a set of 3D radar data with known GB locations $\textbf{x}_k^*$. Fig.~\ref{fig:GBvariation} shows a typical distribution of the GB location differences between neighboring sample positions, i.e. its first difference: $\Delta\textbf{x}_k=d\textbf{x}_k^*/dk=\textbf{x}_k^*-\textbf{x}_{k-1}^*$. The x-axis in Fig.~\ref{fig:GBvariation} represents $\Delta\textbf{x}_k$ measured in image pixels and the y-axis shows its histogram. It shows that the distribution of $\Delta\textbf{x}_k$ fits a Gaussian model fairly well. Therefore, we employ a Gaussian model to describe the state transition (evolutions of the GB locations) in Eq.~\ref{Eq:system}.

%DATA=readMDFFile('C:\Documents and Settings\Li\My Documents\WichmannData\belvoir_metal_data\Metal_Mine_IED_0-3-6_Run1_155CT.mdr');
%myAligndata3D(DATA);

\begin{figure}%MillbrookData\09_28_05\L02\Bad_GPS\L02L_M1_HIGH_g1_F_06.mdr; fileInd=52
	\centering
	$\begin{array}{cc}
		\hspace{-0.0in} \includegraphics[height=1.3in]{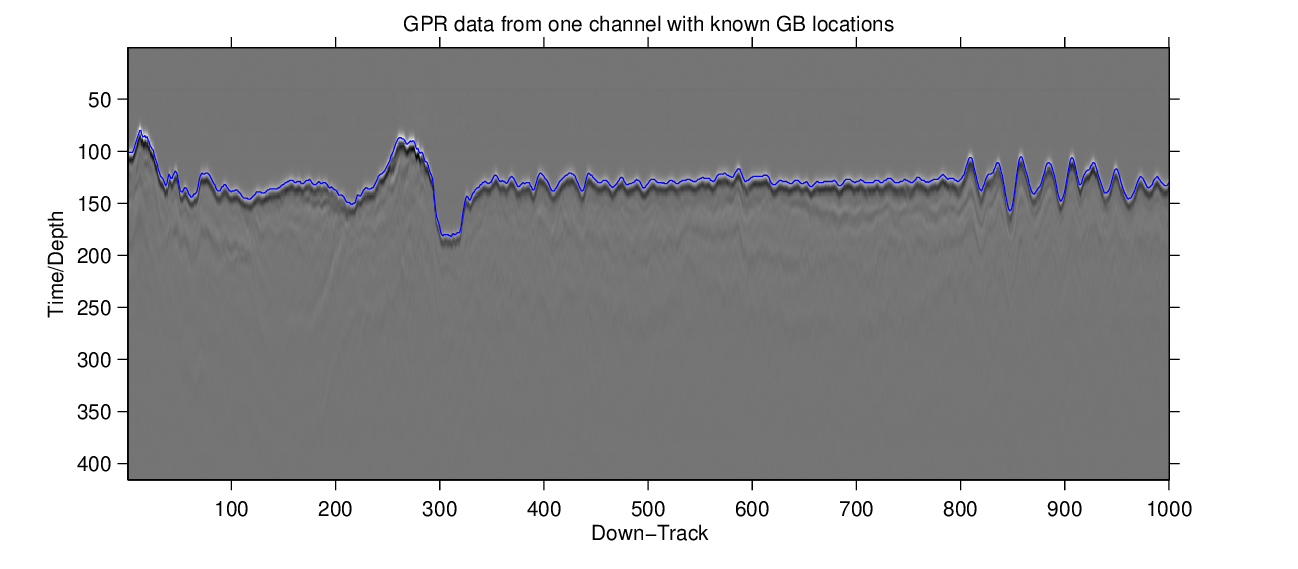} & \\
		\hspace{-0.0in} (a) & \\
		\hspace{-0.2in} \includegraphics[height=1.3in,width=2.6in]{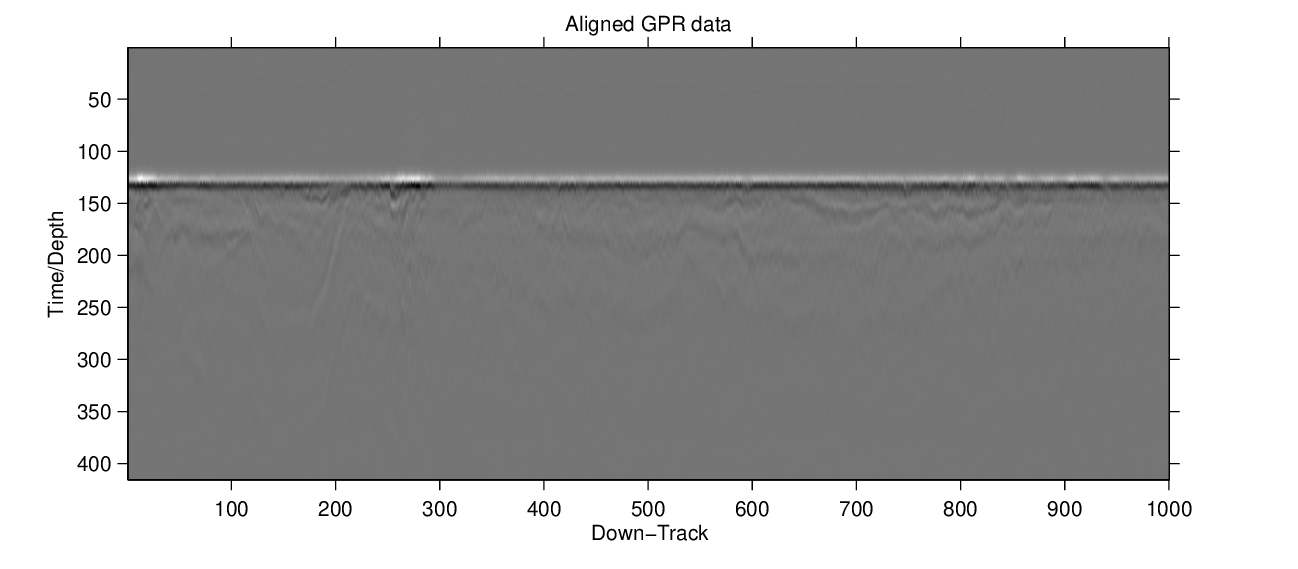} & \hspace{-0.6in} \includegraphics[height=1.3in,width=0.65in]{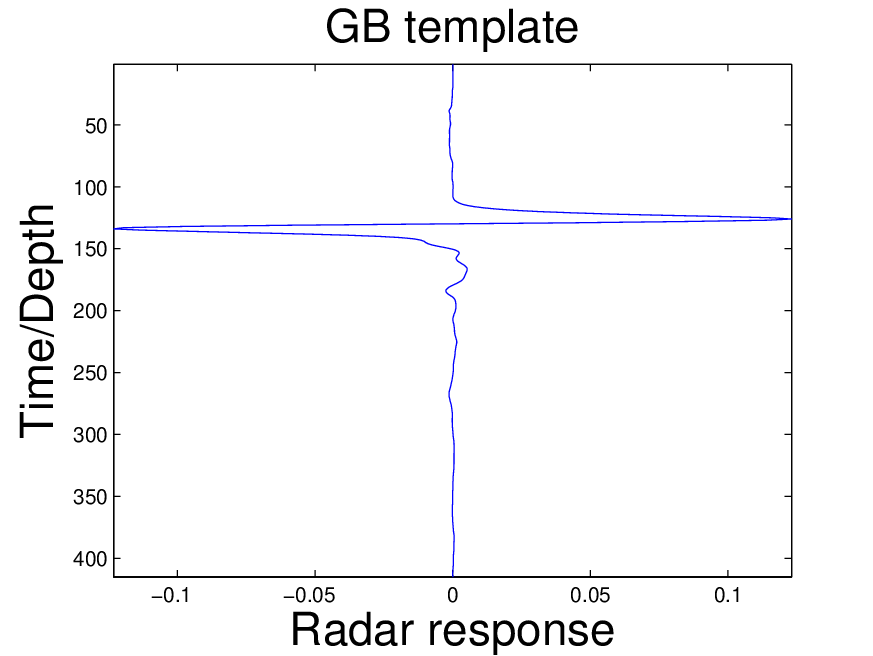}\\
		\hspace{-0.2in} (b) & \hspace{-0.6in} (c)
	\end{array}$
	\caption{Analysis of the observed GB template: (a) GPR data from one channel with known GB locations; (b) Aligned GPR data to place the given GB locations on the same horizontal line; (c) the GB template obtained by averaging the A-scans in (b) along in the down-track direction.}
	\label{fig:GBsignal}
\end{figure}

For the experimental data set with known GB locations (Fig.~\ref{fig:GBsignal}(a)), if we align each vector $\textbf{z}_k$ to make the given GB locations $\textbf{x}_k^*$ lie on the same horizontal line (Fig.~\ref{fig:GBsignal}(b)) and average them along the down-track direction, we obtain a typical GB signature $\textbf{t}_k$ (Fig.~\ref{fig:GBsignal}(c)), where the vertical axis represents the index of the ${N}_d\times{1}$ vector and the horizontal axis gives the magnitude of the GPR response. The GB location can be identified by the sharp change in polarity. At other depths, generally, the signal amplitude is close to zero when no other discontinuities are present. Therefore, given a GB location $\textbf{x}_k$ and the current GB template $\textbf{t}_k$, the observation $\textbf{z}_k$ can be approximated as:
\begin{equation}
\label{Eq:observation}
\textbf{z}_k=\delta(\textbf{x}-\textbf{x}_k)\ast\textbf{t}_k+\textbf{n}_k
\end{equation}

\noindent where $\ast$ denotes the convolution. The length of $\textbf{t}_k$ is determined to contain the main features of the GB signature in Fig.~\ref{fig:GBsignal}(c). Its shape and magnitude may vary for different data sets.

%performanceComparison.m

\subsection{Adaptive template}
\label{Sec:template}

As we observe from Fig.~\ref{fig:GBsignal}, a significant feature of the ground bounce is the large radar response in each A scan. Due to the various conditions such as weather and soil, the shape and magnitude of the peaks may change. In \cite{Wu:AdaptiveGB}, the GB signature is adaptively modeled as a shifted and scaled version of a fixed reference GB. To describe the GB features accurately and to encode them into the tracking process, we represent the GB as an adaptive template, which is expressed as a vector instead of a closed-form function. It is updated with new observations.

\begin{figure}
\centering
	$\begin{array}{cc} %someNotes.m
		\includegraphics[width=1.5in]{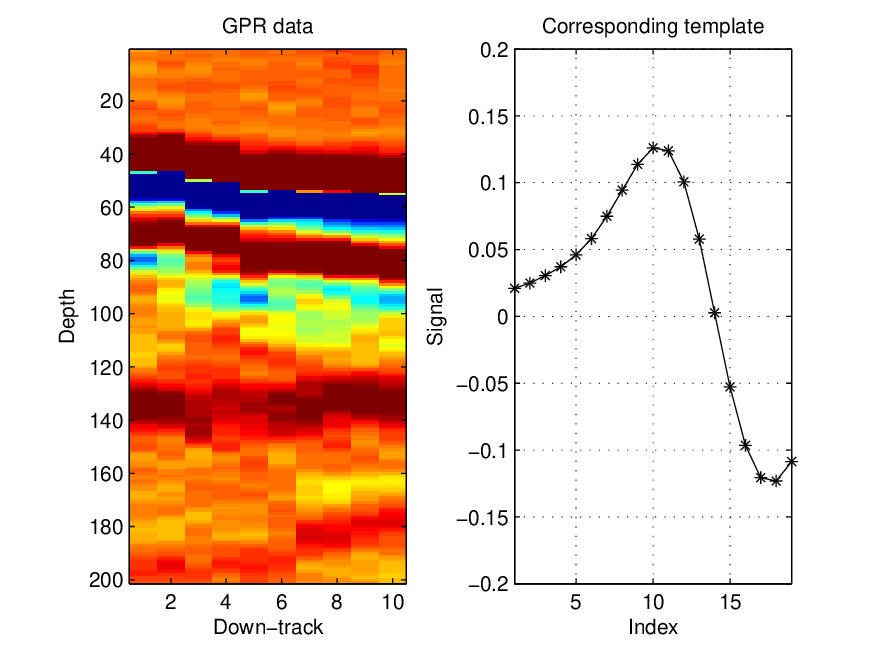} & \includegraphics[width=1.5in]{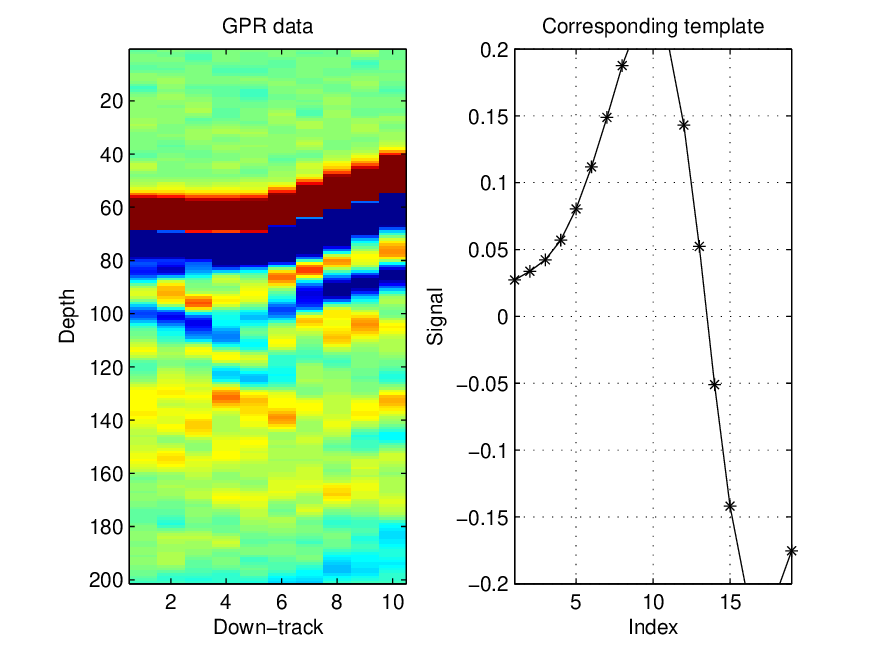}\\
		(a) & (b)\\
		\includegraphics[width=1.5in]{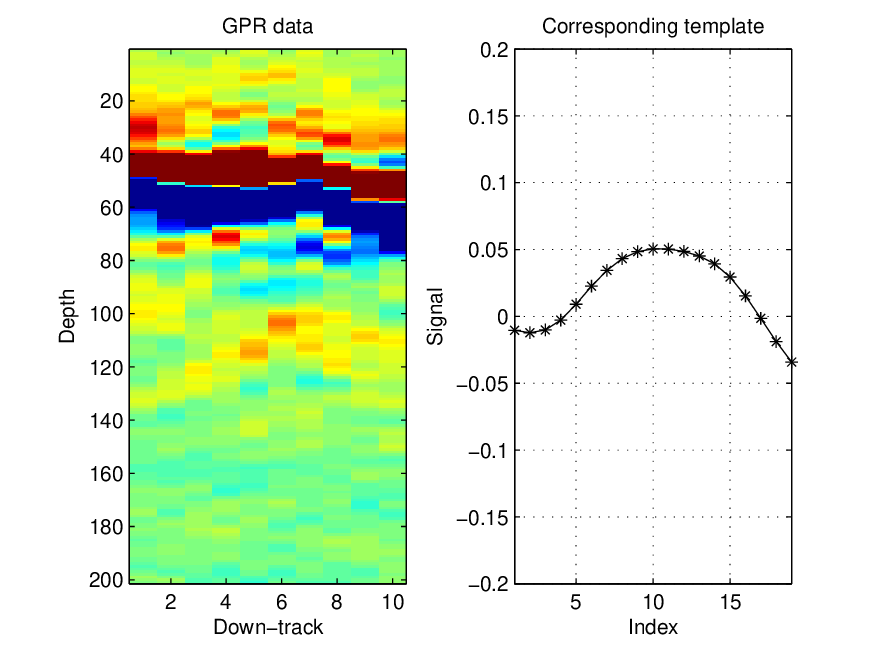} & \includegraphics[width=1.5in]{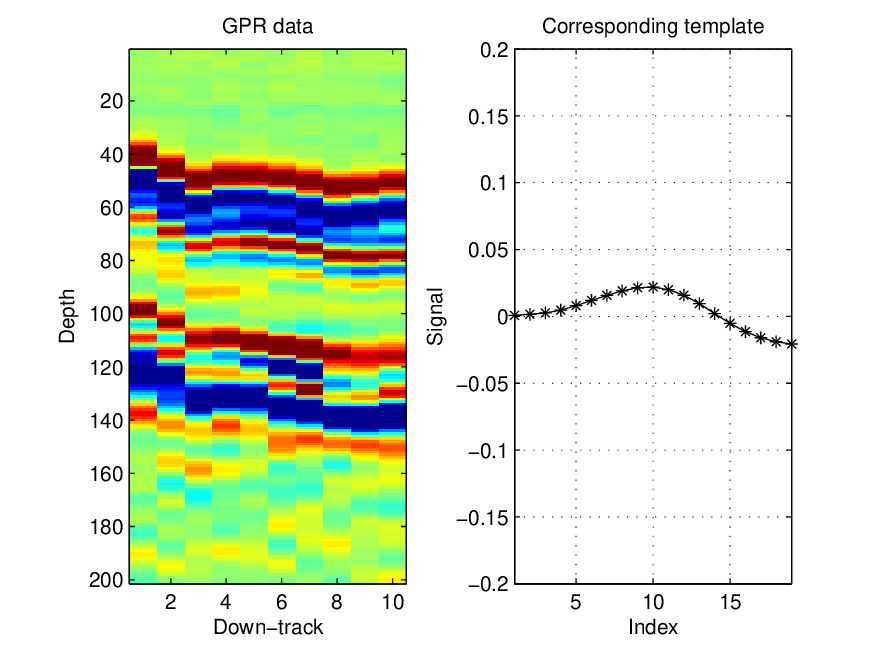}\\
		(c) & (d)%\\
%		\includegraphics[width=1.5in]{template3.eps} & \includegraphics[width=1.5in]{template5.eps}\\
%		(e) & (f)
	\end{array}$
\caption{Various GB templates extracted from the corresponding GPR data. Left panel in each figure shows the GPR data segment; right panel shows the GB template extracted from the data.}
\label{fig:template}
\end{figure}

Fig.~\ref{fig:template} provides several examples of the GB templates denoted as vectors centered around the peak (maximum) with their corresponding GPR data segment. The length of the vector is set to 19 in this case. Fig.~\ref{fig:template} (a)$\sim$(d) show data and GB templates collected from 4 different areas we refer to as lanes (a)$\sim$(d) respectively: Fig.~\ref{fig:template}(a) shows data from a gravel road used for vehicular testing, Fig.~\ref{fig:template}(b) a heavily trafficked gravel access road, Fig.~\ref{fig:template}(c) a grassy area located near lanes (a) and (b), and Fig.~\ref{fig:template}(d) a gravel track covered with 8-12" of snow. Lanes (a)$\sim$(c) are located in the United Kingdom, and lane (d) is located in the northeastern United States. It can be observed that these GB templates have significant differences from each other. Basically, they can be of any shape depending on different data sets without any restrictions enforced by a particular function.

%Millbrook: L01R_M1_HIGH_g1_F_09.mdr; L02L_M1_HIGH_g1_F_06.mdr; L04L_M1_HIGH_g1_F_04.mdr
%CRREL: Lane2_M1_15cm_g1_0_20_Snow.mdr

At time step $k$, once we obtain a ``reliable" GB estimate $\hat{\textbf{x}}_k$, i.e. once the current estimate agrees with the result from the global maximum, we can extract a template $\textbf{t}_c$ of length $2{n}_t+1$ from the current observation $\textbf{z}_k$:
\begin{equation}
\label{Eq:tempExtract}
\textbf{t}_c=[\textbf{z}_k(\hat{\textbf{x}}_k-n_t),\textbf{z}_k(\hat{\textbf{x}}_k-n_t+1),\ldots,\textbf{z}_k(\hat{\textbf{x}}_k+n_t)]
\end{equation}

\noindent where $\textbf{z}_k(m)$ represents the $m^{th}$ element of the current observation vector $\textbf{z}_k$ (A scan) and $\hat{\textbf{x}}_k$ is the estimated one-dimensional state vector (i.e. GB location). Then the current GB template $\textbf{t}_k$ is updated using $\textbf{t}_c$. Otherwise, $\textbf{t}_k$ remains unchanged. In this way, we avoid the template being updated with inaccurate data. The algorithm takes advantage of this extracted template $\textbf{t}_c$ (Fig.~\ref{fig:template}) to encode the features of the GB signature in a more detailed fashion, hence ensuring reliable tracking.

Although the GB template is expected to change adaptively with different data sets and the inhomogeneous ground surface conditions, it must not be influenced significantly by abrupt interference from random clutters. Therefore, we introduce a confidence factor $\alpha_k$ to describe our confidence in the current GB template $\textbf{t}_k$, which is defined as the number of the reliable templates $\textbf{t}_c$ we have obtained up to time step $k$. This definition means that we accumulate confidence as more and more data are processed. The accumulated template $\textbf{t}_k$ is updated accordingly as:
\begin{equation}
\label{Eq:template}
\textbf{t}_k=\frac{\alpha_{k-1}\textbf{t}_{k-1}+\textbf{t}_c}{\alpha_{k-1}+1}=\frac{\alpha_{k-1}\textbf{t}_{k-1}+\textbf{t}_c}{\alpha_k}.
\end{equation}

\noindent It is shown in Eq.~\ref{Eq:tempExtract} that the extraction of a GB template is dependent upon both the observation $\textbf{z}_k$ and the estimate $\hat{\textbf{x}}_k$. The random clutter in the observation $\textbf{z}_k$ is eliminated by averaging $\textbf{t}_c$ from each time step to obtain a stable accumulated $\textbf{t}_k$ (Eq.~\ref{Eq:template}).
%The accuracy of the estimate $\hat{\textbf{x}}_k$ is double checked using the result from the global maximum. If it is a ``reliable" estimate (i.e. once the current estimate agrees with the result from the global maximum), the template $\textbf{t}_k$ is updated according to Eq.~\ref{Eq:template}. 

However, even if the template is updated adaptively, the magnitude of the GB signature under consideration might be slightly different from the current template $\textbf{t}_k$. For example, such changes can occur when the relative distance between the radar and the ground changes. This may cause some errors in the likelihood computation and as a result, the estimated GB locations may slightly deviate from the true peaks in the collected radar data. To eliminate this error, we find the maximum value within a small vicinity of the estimated GB location and take that position as the final estimate. In our implementation, the window size is set to be equal to the length of the template.

% needed in second column of first page if using \pubid
%\pubidadjcol

\subsection{Likelihood computation}
\label{Sec:Likelihood}

The likelihood is used to evaluate the importance of each particle and update its associated weight according to the current observation. It is one of the key calculations for the PF to achieve good performance. To update weights recursively in every iteration according to Eq.~\ref{Eq:weight}, we need to find an appropriate likelihood function for each prior sample $p(\textbf{z}_k|\textbf{x}_k^i)$. With the aforementioned formulation of the observation model (Eq.~\ref{Eq:observation}) and the statistics of the measurement noise $\textbf{n}_k$, the likelihood function can be expressed as follows according to its definition \cite{Gordon:Novel}:
\begin{eqnarray}
\label{Eq:likelihood}
p(\textbf{z}_k|\textbf{x}_k^i) \hspace{-0.15in} & = & \hspace{-0.15in} \int\delta(\textbf{z}_k-h_k(\textbf{x}_k^i,\textbf{n}_k))p_n(\textbf{n}_k)d\textbf{n}_k \nonumber\\
\hspace{-0.15in} & = & \hspace{-0.15in} \int\delta(\textbf{z}_k-\delta(\textbf{x}-\textbf{x}_k^i)\ast\textbf{t}_k-\textbf{n}_k)p_n(\textbf{n}_k)d\textbf{n}_k \nonumber\\
\hspace{-0.15in} & = & \hspace{-0.15in} p_n(\textbf{z}_k-\delta(\textbf{x}-\textbf{x}_k^i)\ast\textbf{t}_k).
\end{eqnarray}

\noindent Since $\textbf{n}_k$ is supposed to obey a zero mean Gaussian distribution $\textbf{n}_k\sim{N}(0,\sigma_n^2)$, it follows that
\begin{equation}
\label{Eq:conditionalPDF}
p(\textbf{z}_k|\textbf{x}_k^i)=\frac{1}{\sqrt{2\pi}\sigma_n}exp\left[-\frac{(\textbf{z}_k-\delta(\textbf{x}-\textbf{x}_k^i)\ast\textbf{t}_k)^2}{2\sigma_n^2}\right].
\end{equation}

To maintain a balance between the focus and the diversity of the particle distribution \cite{Doucet:Book}, the particles should be treated differently at the predict and update stages. At the predict stage, particles are expected to have sufficient diversity based on the prior distribution so that the right state is included in the particle set, which requires a good control on the variance of the process noise $\textbf{v}_k$. It has to be large enough to simulate possible variations of GB locations across different channels/scans, while simultaneously small enough to focus the limited number of the samples within the important regions of the posterior \cite{Wang:Modular}. Actually, the latter goal enforces implicitly the smoothness constraint and is important to achieve an accurate GB tracking. Moreover, in this low-variance case, with an accurate model of the deterministic part of the system dynamics, the efficiency of the particles is improved as well, which means the true continuous distribution can be accurately approximated with only a limited number of discrete particles.

On the other hand, at the update stage, most of the particles should be concentrated in the correct state after the resampling, which provides a good starting point for the prediction in the next iteration. To achieve this goal, the weights should be assigned properly according to the likelihood computation (Eq.~\ref{Eq:weight}). That is, we should not only distinguish those large weights from small ones, but also make their differences significant. Ideally, the likelihood function only assigns a few particles large weights \cite{Jeroen:Influence}. Those particles whose states are far away from the right state are assigned small weights which are usually ignored after the resampling.

\begin{figure}%bootstrap_0621.m(fileInd=52)
\centering
$\begin{array}{c}
\hspace{-0.2in} \includegraphics[width=3.5in,height=1in]{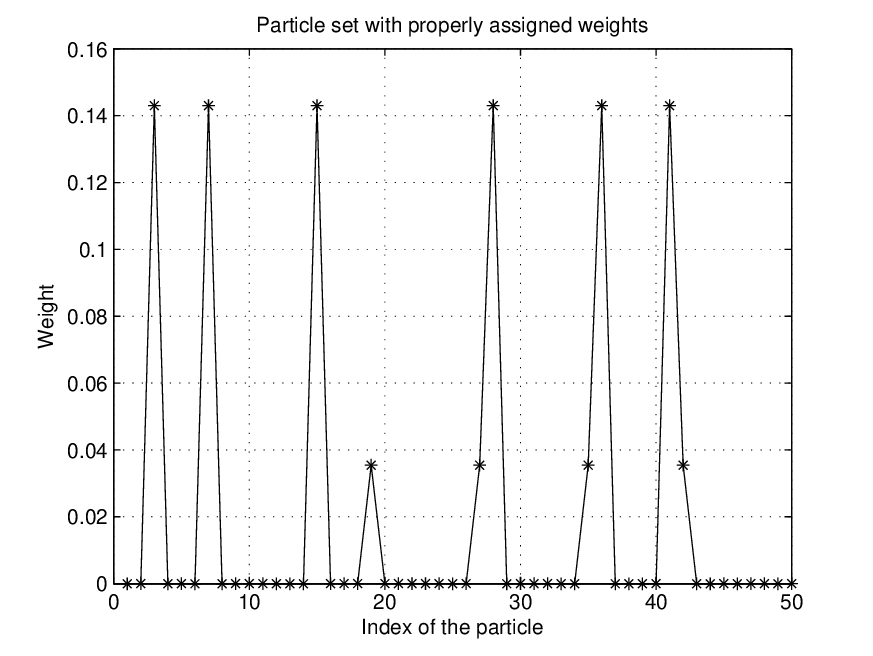} \\
\hspace{-0.2in} \includegraphics[width=3.5in,height=1in]{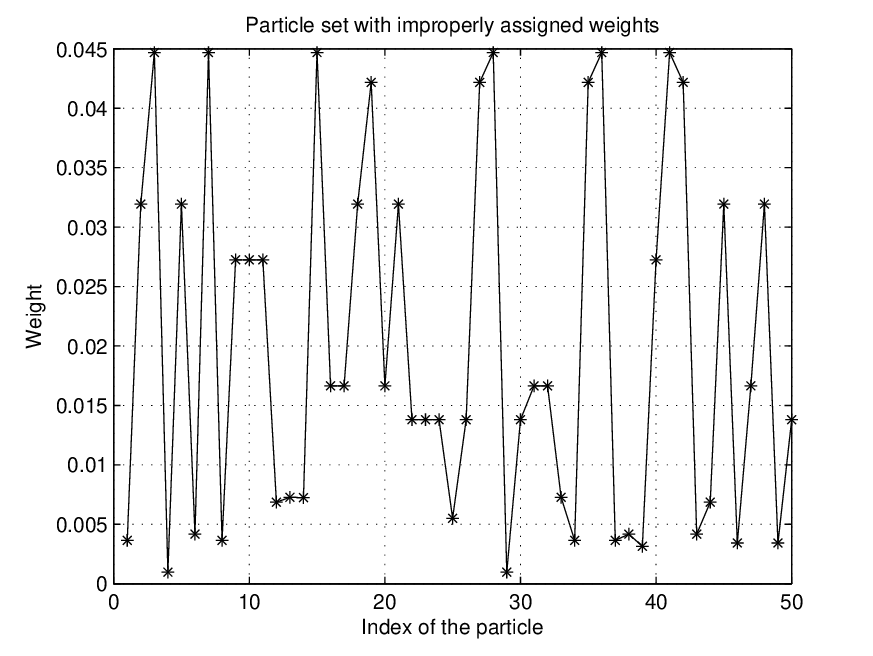}
\end{array}$
\caption{Particle distribution by assigning the appropriate weights. Top panel shows an example of a good weight distribution; bottom panel shows the same particle set with improperly assigned weights.}
\label{fig:weight}
\end{figure}

Fig.~\ref{fig:weight} shows an example of this. In both panels, the x-axis represents the index of the 50 particles and the y-axis shows their assigned weights. The six particles with index 3, 7, 15, 28, 36 and 41 receive the highest weight while the four particles with index 19, 27, 35 and 42 receive the second highest weight. The top panel is an example of a good weight distribution, in which only these 10 ``important" particles will survive after the resampling. However, in the bottom panel, other erroneous particles, whose states are more than 3 pixels away from those ``important" particles, will also exist after the resampling due to their non-neglectable weights. That means those particles are scattered in the state space and wasted in the unimportant regions of the posterior distribution. Therefore, this latter case should be avoided to keep particles of high quality. By concentrating all of the particles around the important regions of the state space, the resolution of the estimate is improved as well.

The likelihood function in Eq.~\ref{Eq:conditionalPDF} can be regarded as a function to evaluate the similarity between the observation $\textbf{z}_k$ and the signature produced by a particle $\delta(\textbf{x}-\textbf{x}_k^i)\ast\textbf{t}_k$ given the current template $\textbf{t}_k$. The term
\begin{equation}
s_i=-\frac{(\textbf{z}_k-\delta(\textbf{x}-\textbf{x}_k^i)\ast\textbf{t}_k)^2}{2\sigma_n^2}
\end{equation}

\noindent acts as a similarity comparison while the exponential function has a nonlinear stretching effect to control the likelihood (weight) differences between the particles. The variance of the observation noise, $\sigma_n$, determines the degree of this stretching and hence the particle distribution (Fig.~\ref{fig:weight}).
%Fig.~\ref{fig:2function} demonstrates the nonlinear stretching effect of the exponential function $y=exp(x)$ by a comparison with the linear function $y=x$. Suppose there are two inputs to the exponential part of the likelihood function $s_1$ and $s_2$, which determine the weights associated with two particles $\textbf{x}_k^1$ and $\textbf{x}_k^2$. By the exponential function, their difference can be ``stretched" from $d_1=\left|s_1-s_2\right|$ to $d'_1$, or similarly, from $d_2$ to $d'_2$ depending on the values of $s_1$ and $s_2$. As shown in Fig.~\ref{fig:2function}, $d_1$ is equal to $d_2$ while after the nonlinear stretching, $d'_2$ is larger than $d'_1$. That is to say, the degree of this stretching effect depends on the input to the exponential function (i.e. $s_1$ and $s_2$), which can be controlled by the variance of the observation noise ($\sigma_n$).
%\begin{figure}
%\centering
%\includegraphics[width=2.5in]{2functions.eps}%someNotes.m
%\caption{Nonlinear stretching effect of the exponential function.}
%\label{fig:2function}
%\end{figure}

Since the weights of all the particles sum up to one, i.e. $\sum_i{w}_k^i=1$, a good particle distribution as the one shown in the top panel of Fig.~\ref{fig:weight} can be obtained if the highest weight in the particle set is large enough. Thus the weights of those unimportant particles are suppressed
%In Fig.~\ref{fig:weight}, the largest weight in the top panel is close to 0.14 so that the weights of those unimportant particles are suppressed. In contrast, the largest weight in the bottom panel is close to 0.045 so that all of the paticles receive comparable weights, which means that particles are scattered in the state space instead of concentrating on the important spot. This situation has to be avoided as we are trying to approximate a continuous distribution with a limited number of discrete particles, which requires
and all of the survived particles convey useful information in the propagation. Experimentally, a reasonable weight for every particle falls within a range of [0,0.2] given the number of particles $N_p$ being 50. In our implementation, $\sigma_n$ is adjusted adaptively to adjust the maximum value of the term $s_i$ in the exponential function to around -0.3 for all of the particles depending on different data sets, i.e. $s_i\leq{-0.3},\mbox{ }i=1,\ldots,N_p$.

To avoid the tracking errors caused by the potential magnitude inconsistency of the estimated template with the current radar signature, a normalization process is applied to both signals ($\textbf{t}_k$ and $\textbf{z}_k$) to ensure that their values fall between 0 and 1 prior to the weight computation. Thus, even if the magnitude of the template $\textbf{t}_k$ slightly deviates from the current GB peak in the observation $\textbf{z}_k$, the right particle will receive the largest weight using the simple similarity comparison of the squared differences ($(\textbf{z}_k-\delta(\textbf{x}-\textbf{x}_k^i)\ast\textbf{t}_k)^2$).

\subsection{Two-way propagation}

\begin{figure}
\centering
\hspace{-0.35in}\includegraphics[width=3.5in,height=2.5in]{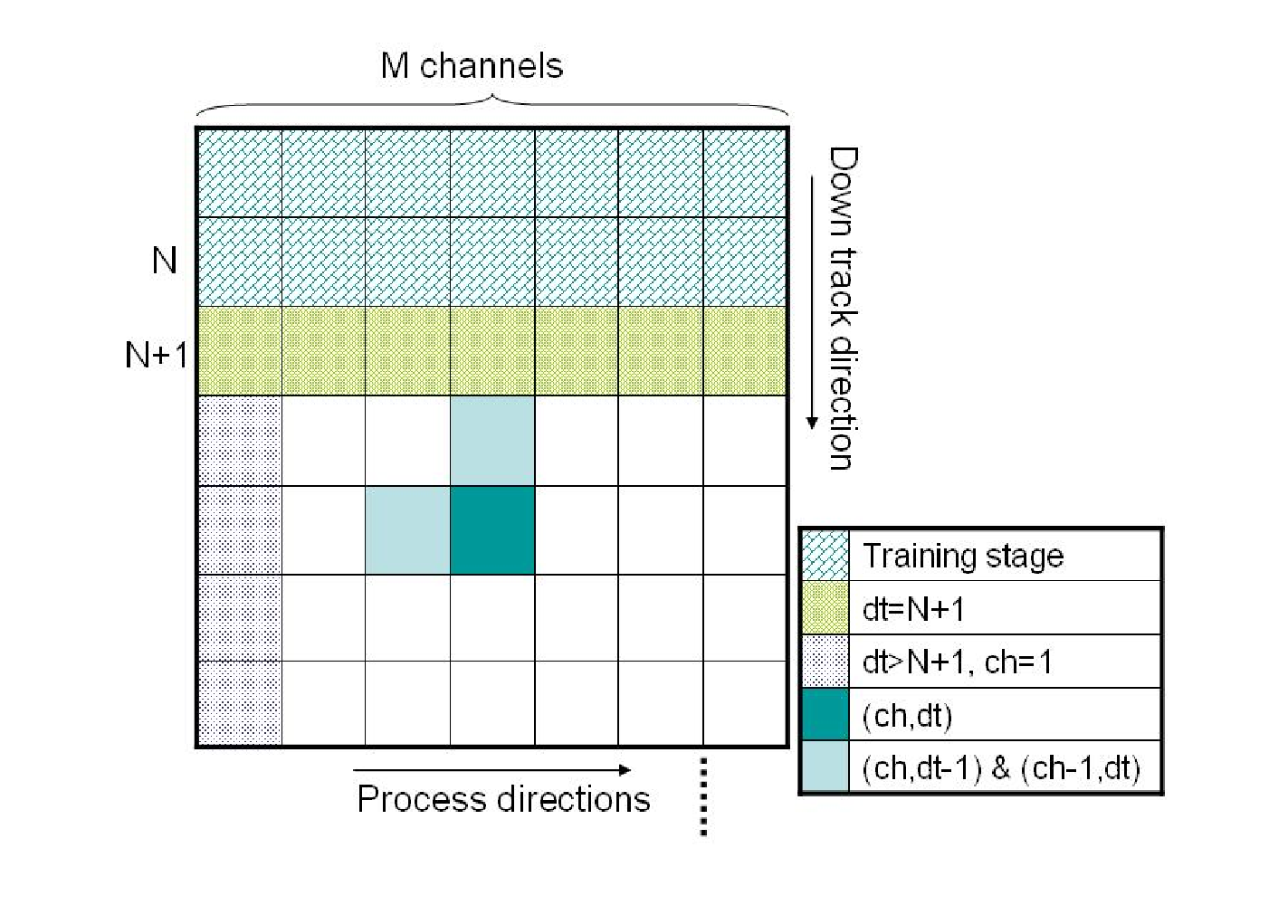}%6_27_2006.ppt
\caption{The data processing order and the two-way propagation.}
\label{fig:process}
\end{figure}

At the prediction stage, samples are drawn according to a prior distribution based on the state-space transition model (Eq.~\ref{Eq:system}). For the first order Markov system, the particle distribution at the current time step is assumed to be only dependent on its most recent history. Specifically, the data are received and processed scan by scan along the down-track direction in the Wichmann/Niitek GPR system. And for each scan, the GB tracking is performed from the first channel to the last channel, which is illustrated in Fig.~\ref{fig:process}. Every DT/XT position is represented as a grid in the figure.

When the first $N$ scans are received and the training stage (Appendix) is completed, the PF is ready to generate output. At the $N+1^{th}$ scan, prior samples are drawn from a Gaussian distribution with the mean being the GB location of its previous scan at the same channel. In the following scans (i.e. $dt>N+1$), there are two ``neighbors" which can be taken as the most recent history for each DT/XT position $(ch,dt)$: the one at the previous scan from the same channel $(ch,dt-1)$ and the one at the same scan from the previous channel $(ch-1,dt)$ (except the first channel). Since the GB locations are assumed to be on a smooth 2D surface, both sets of the particles are propagated through the system transition model (Eq.~\ref{Eq:system}) to obtain two prior distributions for the current position $(ch,dt)$. As there are $N_p$ particles at each of its adjacent positions ($(ch,dt-1)$ and $(ch-1,dt)$), $2\times{N}_p$ particles are produced as the prediction. To keep the number of the particles constant, we draw $N_p$ samples from them with probabilities proportional to their weights, which is similar to the resampling step. For the first channel $(ch=1,dt)$, only particles in the previous scan of that channel $(ch=1,dt-1)$ are propagated to get the prior distribution. To summarize, the propagation is performed as follows:
\begin{eqnarray}
\bar{\textbf{x}}_k^i\hspace{-0.15in} & \sim\hspace{-0.1in} & p(\textbf{x}_k|\textbf{x}_{k-1}) \nonumber\\
\hspace{-0.2in} & \sim\hspace{-0.15in} & \left\{
\begin{array}{cc}
\hspace{-0.05in}N(\textbf{x}_{ch,dt-1},\sigma_v^2), &\hspace{-0.15in} dt=N+1\\
\hspace{-0.05in}p(\textbf{x}_{ch,dt}|\textbf{x}_{ch-1,dt},\textbf{x}_{ch,dt-1}), &\hspace{-0.15in} dt>N+1,ch>1\\
\hspace{-0.05in}p(\textbf{x}_{ch,dt}|\textbf{x}_{ch,dt-1}), &\hspace{-0.15in} dt>N+1,ch=1
\end{array} \right.
\end{eqnarray}

\noindent where $ch$ and $dt$ ($ch=1,\ldots,N_{CH}$, $dt=N+1,\ldots,N_{DT}$) are the cross-track and down-track indexes, respectively. The so called ``time step" $k$ is related with $ch$ and $dt$ as:
\begin{equation}
k=(dt-1)\times{N}_{CH}+ch.
\end{equation}

\begin{figure*}%predictstates_0621.m
\centerline{\subfigure[$\textbf{x}_{ch-1,dt}$]{\includegraphics[width=2in]{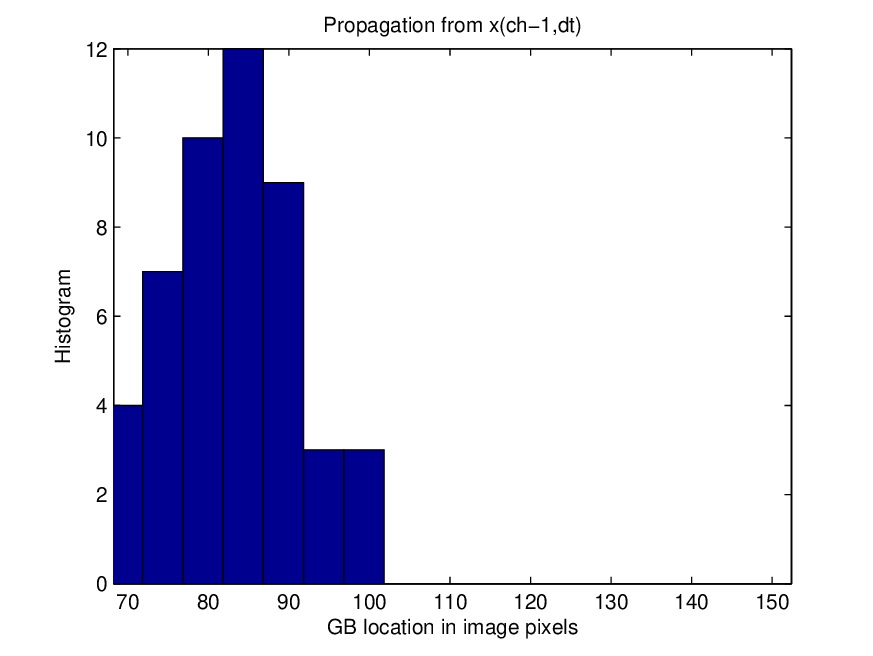}
\label{fig:hist1}}
\hfil
\subfigure[$\textbf{x}_{ch,dt-1}$]{\includegraphics[width=2in]{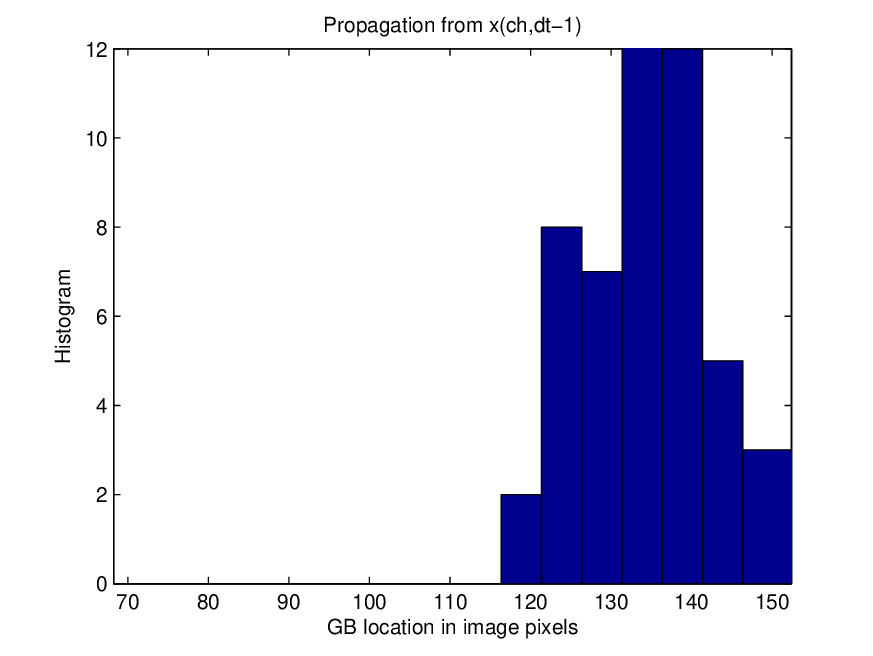}
\label{fig:hist2}}
\hfil
\subfigure[$\textbf{x}_{ch,dt}$]{\includegraphics[width=2in]{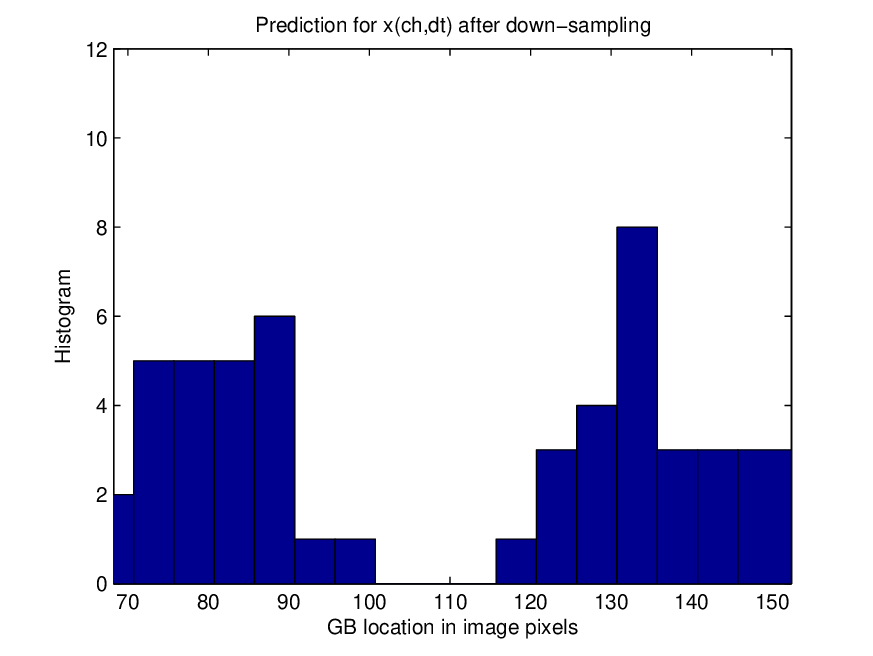}
\label{fig:hist3}}}
\caption{One example of a multimodal prior distribution predicted by the two-way propagation.}
\label{fig:hist}
\end{figure*}

Fig.~\ref{fig:hist} shows one example of the proposal distribution derived from the two-way propagation. They are the particle distributions at three adjacent positions ($\textbf{x}_{ch-1,dt}$, $\textbf{x}_{ch,dt-1}$ and $\textbf{x}_{ch,dt}$), where the x-axis represents the GB locations and the y-axis shows their distributions. $2\times{N}_p$ particles are drawn by propagation from the two adjacent locations (Fig.~\ref{fig:hist1} and Fig.~\ref{fig:hist2}) according to the system transition model (Eq.~\ref{Eq:system}).
%which improves the GB tracking accuracy by implicitly incorporating the smoothness constraint of the ground into the prediction stage.
These $2\times{N}_p$ particles are down-sampled to keep the number of the particles to be constant at ${N}_p$. The resulting prediction is shown in Fig.~\ref{fig:hist3}. This is a special example that the priors at its two adjacent locations $\textbf{x}_{ch-1,dt}$ and $\textbf{x}_{ch,dt-1}$ do not agree with each other, as observed from the x-axis of Fig.~\ref{fig:hist1} and Fig.~\ref{fig:hist2}. That means they suggest different ``proposals" for the possible GB locations at the current position ($\textbf{x}_{ch,dt}$). Two-way propagation takes both proposals into account to ensure the prior samples are drawn from the right places in the state space. Specifically, a multimodal distribution is derived as the prior in this example, which includes two peaks in Fig.~\ref{fig:hist3}. This case is more convenient for the PF to handle than for the KF, as it does not obey a simple Gaussian distribution. However, in most cases, these two neighbors usually suggest similar prior distributions.

\section{Comparison of KF and PF formulations}
\label{Sec:comTable}

A comparison of the KF and the PF formulations is summarized in Table~\ref{table:KF_PF}.

\begin{table}
\renewcommand{\arraystretch}{1.3}
\caption{Comparison of KF and PF formulations}
\label{table:KF_PF}
\centering
\begin{tabular}{|c||c|c|}
\hline
Algorithm & KF & PF\\
\hline\hline
State $\textbf{x}_k$ & $[GB_{ch}^k\mbox{ }GB_{ch-1}^k\mbox{ }GB_{ch+1}^k\mbox{ }{GB'}_{ch}^k\mbox{ }{GB'}_{ch-1}^k\mbox{ }{GB'}_{ch+1}^k]^T$ & $\{\textbf{x}_k^i,w_k^i\}_{i=1}^{N_p}$\\
\hline
Observation $\textbf{z}_k$ & GB tracking result from the GM: $GB_{max}$ & A-scan in the 3D GPR signature\\
\hline
State Transition & $\textbf{x}_k=f_k(\textbf{x}_{k-1},\textbf{v}_k)=\textbf{F}\textbf{x}_{k-1}+\textbf{v}_k$ & $\textbf{x}_k^i\sim{p}(\textbf{x}_k|\textbf{x}_{k-1})$\\
\hline
Observation Model & $\textbf{z}_k=h_k(\textbf{x}_k,\textbf{n}_k)=\textbf{H}\textbf{x}_k+\textbf{n}_k$ & $\textbf{z}_k=\delta(\textbf{x}-\textbf{x}_k)\ast\textbf{t}_k+\textbf{n}_k$\\
\hline
Adaptive Parameters & Observation noise covariance & Template and Observation noise covariance\\
\hline
%Process Noise & $\textbf{v}_k\sim{N}(0,\textbf{Q}_k),\textbf{Q}_k$ is low and constant & $\textbf{v}_k\sim{N}(0,\sigma_v^2)$\\
%\hline
%Observation Noise & $\textbf{n}_k\sim{N}(0,\textbf{R}_k),\textbf{R}_k$ is high and variable & $\textbf{n}_k\sim{N}(0,\sigma_n^2)$\\
%\hline
\end{tabular}
\end{table}

\section{Experimental results}
\label{Sec:experiments}

\subsection{Data collection and ground truth}

To evaluate the robustness of the proposed algorithms to different road and weather conditions, we tested them with a wide variety of data sets, which include data collected in clear weather and cold weather, damp soil and dry soil, sandy, gravel and asphalt ground surfaces. The data were collected at government managed test sites in both the eastern and western US and in the UK. Both plastic and metal-cased anti-tank landmines were buried in the lanes. A detailed data description is provided in Table~\ref{table:DATA}. By comparing the results from all of the available algorithms, and determining the GB tracking errors when the trackers differed substantially, we manually defined a GB ``ground truth" and use it to evaluate the performance of the different algorithms.

\begin{table*}
\renewcommand{\arraystretch}{1.3}
\caption{Descriptions of the experimental data sets}
\label{table:DATA}
\centering
\begin{tabular}{|c||c|c|c|c|}
\hline
Data & Scans & Square Meters Covered & Targets & Descriptions\\
\hline\hline
T1 & 61 319 & 3679.1 & 38 & Standard test lanes\\%Lane3 and Lane4
\hline
T2 & 36 512 & 2190.7 & 30 & Standard test lanes\\%YUMA: Lane51 and Lane52
\hline
T3 & 69 990 & 4199.4 & 0 & off road clutter lanes\\%APG
\hline
T3a & 5005 & 300.3 & 44 & Standard test lanes (with snow)\\%CARCASS: Lane1,2,4,6,8,10,12
\hline
T4 & 36 381 & 2182.9 & 53 & Standard test lanes\\%Aphill: Lane3,4,13,14
\hline
T4a & 25 977 & 1558.6 & 103 & Standard test lanes (with surface mines)\\%Lane4 and Lane19
\hline
T5 & 10 944 & 656.64 & 24 & Standard test lanes on asphalt (with surface mines)\\%Run1,2,4
\hline
T6 & 26 459 & 1587.5 & 61 & Roads (with surface mines)\\%Millbrook: Lane1,2,4
\hline
T6a & 84 907 & 5094.4 & 0 & Roads\\%Millbrook: Lane8,9,10
\hline
\end{tabular}
\end{table*}

\subsection{Comparison of the GB tracking performances using data collected in cold weather}
\label{Sec:snow}

In benign conditions, such as a clear weather and unobstructed ground, most GB trackers provide similar results. The advantages of a robust tracker can be observed in more difficult scenarios, such as data taken under snow cover or with other heavy surface clutters.

\begin{figure}%performanceComparison.m
\centering
$\begin{array}{c}
\hspace{-0.2in}\includegraphics[width=3.5in,height=2in]{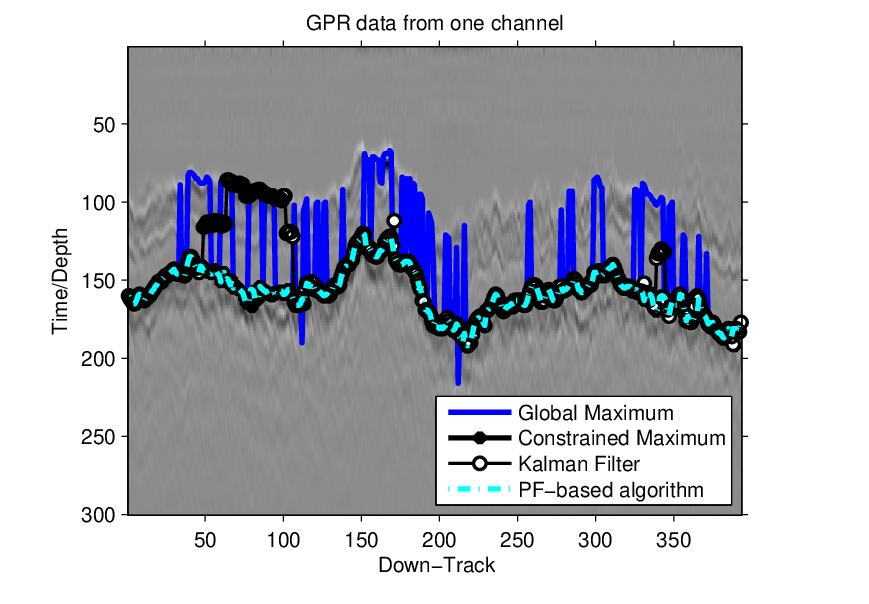}\\%compare_snow.eps
(a)\\
\hspace{-0.2in}\includegraphics[width=3in]{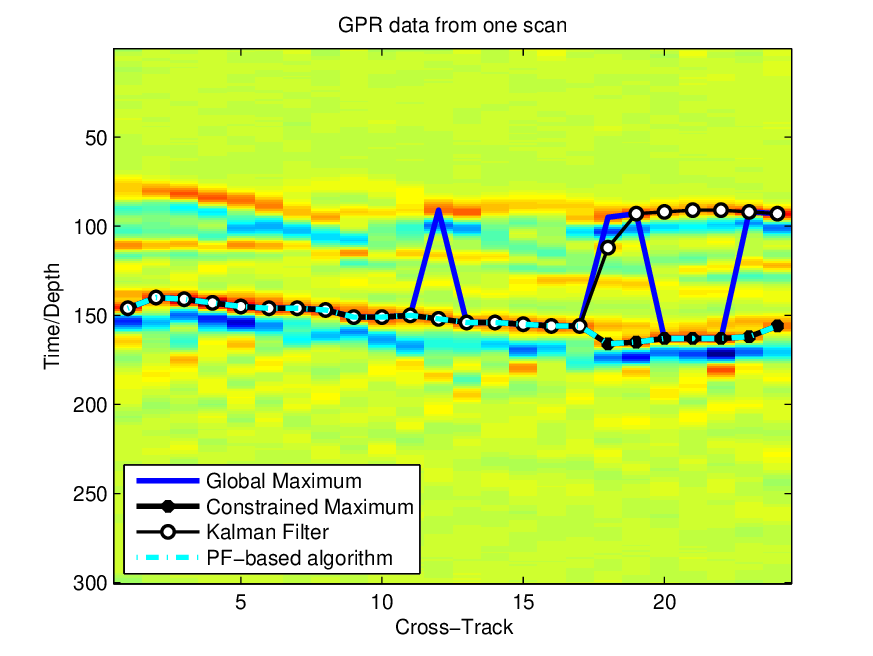}\\
(b)
\end{array}$
\caption{Comparison of the GB tracking performance using data taken under snow cover: (a) One channel; (b) One scan.}
\label{fig:compare_snow}
\end{figure}

The performances of the KF and the PF algorithms are demonstrated by comparisons with the results from the global maximum and the constrained maximum. One B scan is shown in Fig.~\ref{fig:compare_snow}(a), in which the x-axis represents the down-track direction and the y-axis represents the time/depth. As this data set was collected in cold weather, there are layers of snow covering the ground, which makes the GB tracking problem much more challenging. As a result, the global maximum (dark solid line) produced several errors, where the estimated GB ``jumps" from the ground/snow interface to the snow/air interface instead of staying at the ground/snow interface. The KF (black solid line with white circle markers) alleviated the errors to some extent. But at locations where the global maximum is erroneously tracking the snow for an extended time, the KF loses track. In contrast, the PF approach (bright dotted line) works fairly well in spite of the difficult conditions. The number of particles, $N_p$, is 50 in this case. The estimated GB locations are exactly the same as those from the global maximum in most positions while the abrupt ``jumps" are eliminated.
%The only error occurred in the initial part of the data set where the output is actually produced by the global maximum as the PF is still in the initial training stage.
With an appropriate size for the search window according to the specific data set, the constrained maximum (black solid line with dots) produced similar results as the PF-based algorithm in this example. Similar results are shown from one scan in Fig.~\ref{fig:compare_snow}(b), in which the x-axis represents the cross-track index and the y-axis represents the time/depth. To show the GB tracking performances clearly, we only plot the first 300 samples along the time/depth dimension in Fig.~\ref{fig:compare_snow}.

\begin{figure}%CARCASS\Lane1_M1_15cm_g1_0_20_Snow_no_water.mdr
	\centering
	$\begin{array}{c}
		\includegraphics[height=1.8in]{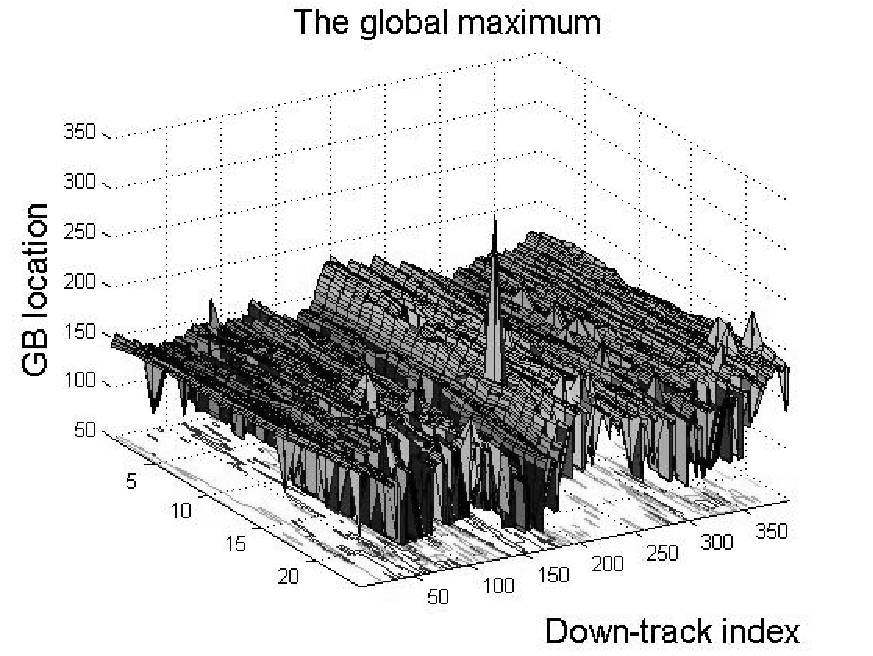}\\%someNotes.m
		(a)\\
		\includegraphics[height=1.8in]{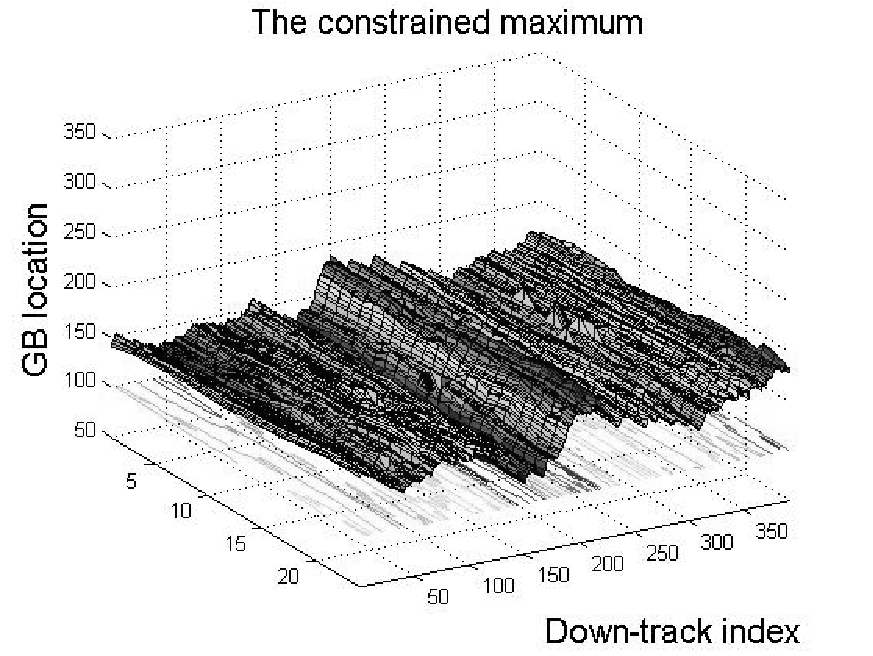}\\
		(b)\\
		\includegraphics[height=1.8in]{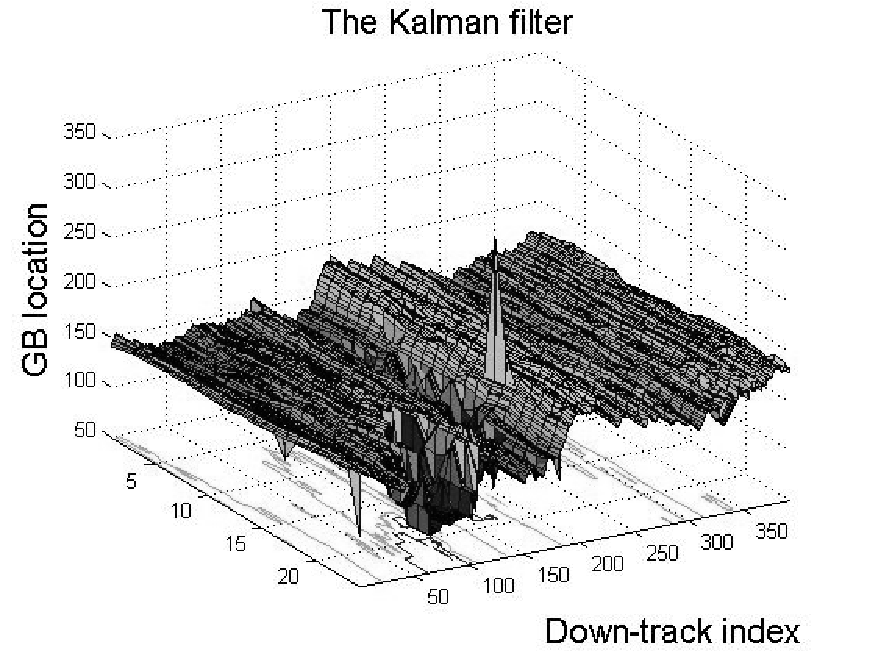}\\
		(c)\\
		\includegraphics[height=1.8in]{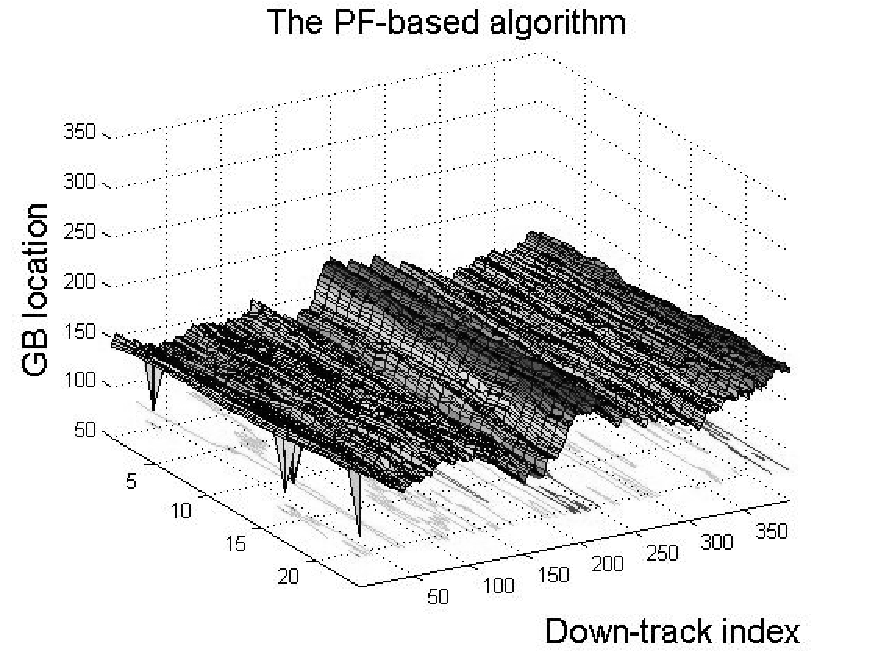}\\
		(d)
	\end{array}$
	\caption{GB tracking results in all 24 channels using data taken under snow cover: (a) The global maximum; (b) The constrained maximum; (c) The Kalman filter; (d) The PF-based algorithm.}
	\label{fig:3Drendering}
\end{figure}

To demonstrate the overall performances across all 24 channels, we plot the GB tracking results as 3D surfaces in Fig.~\ref{fig:3Drendering}. Four approaches are compared here. As expected, there are many errors in the global maximum estimate (Fig.~\ref{fig:3Drendering}(a)), since the GB locations are not always consistent with the maximum responses in each A scan. The constrained maximum (Fig.~\ref{fig:3Drendering}(b)) gives much better results when an size of the search window fits the specific data set. Compared with the KF (Fig.~\ref{fig:3Drendering}(c)), the PF-based algorithm (Fig.~\ref{fig:3Drendering}(d)) has better performance with only a few errors in evidence during the initial training stage. Its accuracy can also be observed from the contour plots provided beneath the 3D surfaces.
%Since there is no general criteria to determine the window size, this is not feasible in real applications.

%GB_callGlobalMaxGBaligningMay01,GB_callLocalMaxGBaligningJuly20
%GB_callKalmanLockGBaligningMay03,GB_callLiGBaligningNewPF50July19

\subsection{Performance comparisons with the GB ground truth}
\label{Sec:GBcomparison}

Since we have located the correct pixel associated with the ground bounce for every DT/XT scan for the data listed in Table~\ref{table:DATA}, we can evaluate the performance of the different algorithms by computing the biases and variances of the different GB tracking results. The comparison is summarized in Table~\ref{table:Bias_Var}. All of the GB trackers produce similar bias values. Therefore, the performances are mainly assessed based on the magnitudes of the variances. As a variety of data sets are considered together, the results for the constrained maximum indicate its sensitivity to a proper window size. The PF-based algorithm outperforms the others with the minimum variance and a similar bias.
%The same conclusions can be drawn from the histogram in Fig.~\ref{fig:bias_var}.

%GB_callGlobalMaxGBaligningMay01,GB_callLocalMaxGBaligningAdaWinDec07
%(GB_callLocalMaxGBaligningadaWinNov27,GB_callLocalMaxGBaligningJuly20)
%GB_callKalmanLockGBaligningMay03,GB_callLiGBaligningNewPF13Dec20
%those workspaces were saved in evaluateAlgorithms.m
%DUKEGroundBounceMaster_v3.m
%C:\Program Files\MATLAB\R2007a\work\niitek\D_GBMv3
\begin{table}
\renewcommand{\arraystretch}{1.3}
\caption{GB tracking performance comparisons with the ground truth data}
\label{table:Bias_Var}
\centering
\begin{tabular}{|c||c|c|c|c|}
\hline
Algorithm & Global & Constrained & Kalman & PF-based\\
 & Maximum & Maximum & Filter & Algorithm\\
\hline\hline
%Bias & 0.003 474 & 0.004 441 & 0.045 322 & -0.012 485\\
%\hline
%Variance & 2.371 0 & 7.956 7 & 1.947 0 & 0.758 3\\
%\hline
Bias & 0.003 474 & 0.027 007 & 0.045 322 & -0.012 485\\
\hline
Variance & 2.371 0 & 4.340 4 & 1.947 0 & 0.758 3\\
\hline
\end{tabular}
\end{table}

%\begin{figure}
%\centering
%\includegraphics[width=3.2in,height=1.8in]{bias_var.eps}
%\caption{Bias and variance of different GB trackers compared with the ground truth.}
%\label{fig:bias_var}
%\end{figure}

\subsection{Performance comparisons using ROC curves}

\begin{figure}%NumberOfSquareMeters.m
	\centering
	$\begin{array}{c}
		\includegraphics[height=2.5in]{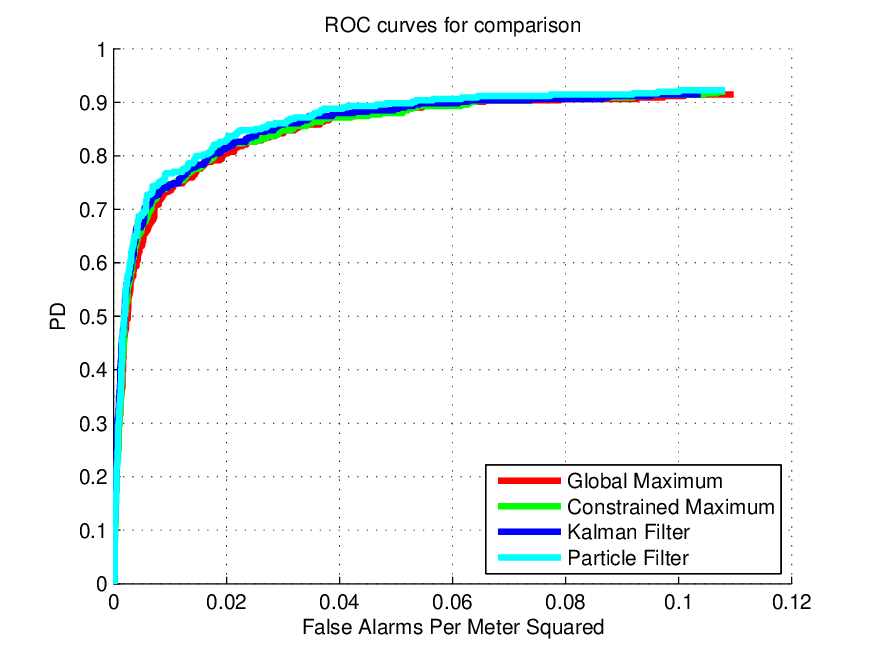}\\
		(a)\\
		\includegraphics[height=2.5in]{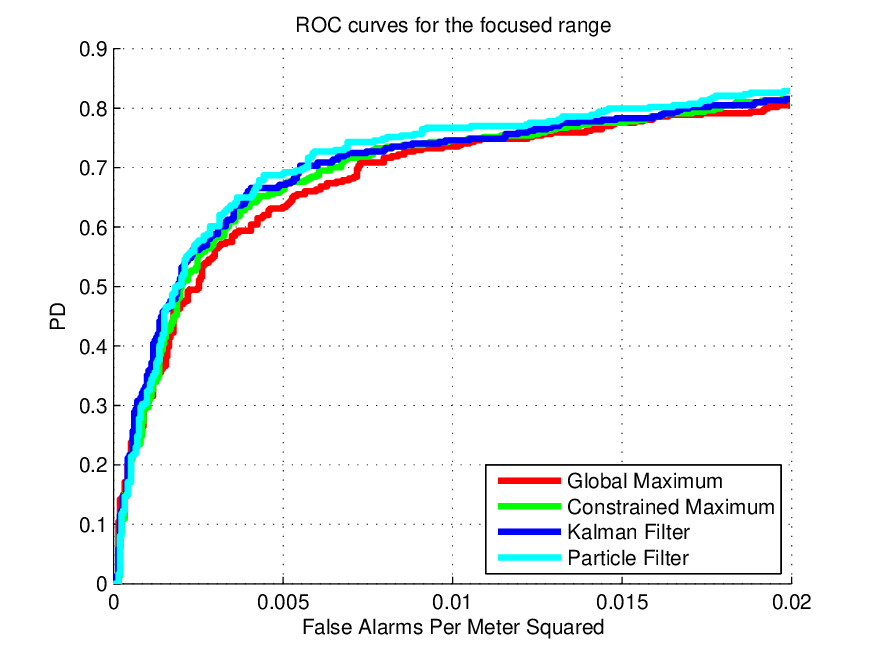}\\
		(b)
	\end{array}$
	\caption{ROC curve comparisons by evaluating different GB tracking methods within the pre-screening algorithm: (a) ROC curves; (b) ROCs for FAR: $0\sim 0.02$.}
	\label{fig:ROC}
\end{figure}

To demonstrate the usefulness of a good GB tracking algorithm in improving the landmine detection accuracy, we provide the different GB tracking outputs to the pre-screening algorithm \cite{Peter:feature}. The landmine detection results are presented in terms of Receiver Operating Characteristic (ROC) curves in Fig.~\ref{fig:ROC}, which aggregates scores over all of the test lanes spanning 21 450 square meters. (Scores were generated using the landmine ground truth provided by the government sponsor.) The x-axis represents the false alarms per meter squared (FAR: false alarm rate) while the y-axis plots the probability of detection (PD). Fig.~\ref{fig:ROC}(b) shows the details of the ROC curves in the focused FAR range of $(0\sim 0.02 FA/m^2)$. It is worth noting that the date listed in Table~\ref{table:DATA} are extremely difficult data for landmine detection. They are used in our experiments to test the behaviors of the GB trackers with only the pre-screener. Under normal conditions, much better ROC curves can be obtained with more complicated feature-based discrimination algorithms following the pre-screener.

%NumberOfSquareMeters.m <- getModeList.m <- generateELEall2.m <- produceROCcurves.m <- (DUKEGroundBounceMaster_v3.m) <- evaluateAlgorithms.m <- To obtain the required ROC curves
%GB_callGlobalMaxGBaligningMay01,GB_callLocalMaxGBaligningAdaWinDec07
%(GB_callLocalMaxGBaligningadaWinNov27,GB_callLocalMaxGBaligningJuly20)
%GB_callKalmanLockGBaligningMay03,GB_callLiGBaligningNewPF13Dec20

%{'GlobalMaxGBalignMay01','LocalMaxGBalignJuly20',...
%'KalmanLockGBalignJuly19','LiGBalignNewPF50July19'}

As in the previous sections, four GB trackers are compared. From Fig.~\ref{fig:ROC}, the performances rank from the best to the worst as: the PF-based algorithm, the KF, the constrained maximum and the global maximum. The KF and the PF outperform the global maximum and the constrained maximum under exactly the same experimental environments.
%which is consistent with the results in Section~\ref{Sec:GBcomparison}. It demonstrates that accurate GB tracking contributes to improved landmine detection.

% Reminder: the "draftcls" or "draftclsnofoot", not "draft", class option
% should be used if it is desired that the figures are to be displayed while
% in draft mode.

% An example of a floating figure using the graphicx package.
% Note that \label must occur AFTER (or within) \caption.
% For figures, \caption should occur after the \includegraphics.
%
% An example of a double column floating figure using two subfigures.
% (The subfigure.sty package must be loaded for this to work.)
% The subfigure \label commands are set within each subfigure command, the
% \label for the overall fgure must come after \caption.
% \hfil must be used as a separator to get equal spacing
%
%\begin{figure*}
%\centerline{\subfigure[Case I]{\includegraphics[width=2.5in]{subfigcase1}
% where an .eps filename suffix will be assumed under latex, 
% and a .pdf suffix will be assumed for pdflatex
%\hfil
%\subfigure[Case II]{\includegraphics[width=2.5in]{subfigcase2}
% where an .eps filename suffix will be assumed under latex, 
% and a .pdf suffix will be assumed for pdflatex
%\label{fig_second_case}}}
%\caption{Simulation results}
%\label{fig_sim}
%\end{figure*}

To evaluate the GB tracking performances quantitively, we compute the areas under the ROC curves (AUC), which is defined as $AUC=\int{PD}d(FAR)$. Clearly, the larger the AUC, the more separable the landmines and the clutters, i.e. the better the performance. The results are listed in Table~\ref{table:AUC}, which are the AUC in the focused FAR range of $0\sim0.02$ (i.e. $AUC=\int_0^{0.02}{PD}d(FAR)$). The same conclusions are observed, which show that both the KF and the PF GB trackers improved the performance of a prescreener for landmine detection.

\begin{table}
\renewcommand{\arraystretch}{1.3}
\caption{GB tracking performance comparisons with the AUC}
\label{table:AUC}
\centering
\begin{tabular}{|c||c|c|c|c|}
\hline
Algorithm & Global & Constrained & Kalman & PF-based\\
 & Maximum & Maximum & Filter & Algorithm\\
\hline
%AUC & 0.8783 & 0.8791 & 0.8817 & 0.8895\\
%\hline
%AUC: ($0\sim0.02$) & 0.0137 & 0.0138 & 0.0141 & 0.0143\\
%\hline
%
%AUC & 0.8494 & 0.8514 & 0.8532 & 0.8640\\
%\hline
AUC: ($0\sim0.02$) & 0.013 5 & 0.013 7 & 0.013 9 & 0.014 1\\
\hline
\end{tabular}
\end{table}

% An example of a floating table. Note that, for IEEE style tables, the 
% \caption command should come BEFORE the table. Table text will default to
% \footnotesize as IEEE normally uses this smaller font for tables.
% The \label must come after \caption as always.
%
%\begin{table}
%% increase table row spacing, adjust to taste
%\renewcommand{\arraystretch}{1.3}
%\caption{An Example of a Table}
%\label{table_example}
%\centering
%% Some packages, such as MDW tools, offer better commands for making tables
%% than the plain LaTeX2e tabular which is used here.
%\begin{tabular}{|c||c|}
%\hline
%One & Two\\
%\hline
%Three & Four\\
%\hline
%\end{tabular}
%\end{table}

\section{Conclusions}
\label{Sec:conclusion}

Different GB trackers are proposed and compared for a landmine detection system. By tracking and removing the GB signatures properly from the 3D data collected by ground penetrating radar, the performance of subsequent landmine detection algorithms is improved. Within a PF framework, the GB tracking is taken as an estimation problem in a dynamic state-space system and the GB location at each DT/XT position is the hidden state to be estimated. The GB signature is described as an adaptive template to accommodate different ground and weather conditions, which is updated upon the receipt of the new observations. This formulation provides more information in describing the GB features in addition to the maximum response in GPR data, which is the only feature in the other GB trackers considered.
%A training stage is involved in forming the initial template and setting the appropriate parameter for the likelihood computation. By predicting the prior distribution from the GB locations of its two adjacent scans/channels, the algorithm takes the smoothness constraint of the ground into consideration instead of searching for GB locations at each DT/XT position independently.
%
Since the state vector is formed as the one dimensional GB location, a limited number of discrete particles is sufficient to approximate its posterior distribution in the state space and particle filtering works efficiently in this scenario with a reasonable computational intensity. Compared with other computationally simple trackers or the tracker using a linear KF, the PF-based algorithm provides a reliable GB tracking at the cost of some increase in the computation time. Its performance is verified both by the GB tracking accuracy and by its usefulness in improving the landmine detection using real GPR data in our landmine detection system. Due to its robustness and flexibility, the algorithm may be extended to other clutter reduction applications in the landmine detection project, which remains to be further considered.

\appendix[PF Training Stage for Parameter Settings]
% or

Compared with the global maximum tracker, the PF provides the flexibility to take many factors into consideration in the GB tracking process. For example, the GB locations in the adjacent DT/XT positions remain smooth (we model the dynamics of the GB location as a first order Markov process) and the GB signature can be of any shape depending on the specific road and weather conditions. However, one disadvantage is that there are a number of parameters which have to be adjusted properly to accommodate the different data sets, such as the constant $\sigma_n$ used to control the focus and the diversity of the particle distributions in Section~\ref{Sec:Likelihood}. In contrast, the global maximum has the advantage of simplicity. Furthermore, in most benign cases, the true GB locations are exactly consistent with the results from the global maximum. Therefore, the global maximum is adopted to provide a kind of ``reference" in a training stage to set the parameters and initiate the PF.

At the training stage, which includes the first $N$ scans of the 3D GPR data, the main tasks are to form the initial GB template $\textbf{t}_k$ and determine the constant $\sigma_n$ in the likelihood function of Eq.~\ref{Eq:conditionalPDF} for each data set. Let the output of the GB tracking be the result from the global maximum temporarily. At every time step, we extract one $\textbf{t}_c$ (Eq.~\ref{Eq:tempExtract}). By averaging all of these $\textbf{t}_c$ together, an initial template $\textbf{t}_k$ is formed. The confidence factor at this point is: $\alpha_{ini}=N_{CH}\times{N}$. The template $\textbf{t}_k$ is then updated constantly throughout the process as mentioned in Section~\ref{Sec:template}. It is worth noting that the template $\textbf{t}_k$ is not updated according to Eq.~\ref{Eq:template} from the very first scan. Only after the training stage, which assumes that the algorithm gains enough knowledge based on the results from the global maximum, is the PF expected to provide stable output.

The constant $\sigma_n$, which is used in the likelihood computation (Eq.~\ref{Eq:conditionalPDF}), is a parameter to be determined in this stage as well. As discussed in Section~\ref{Sec:Likelihood}, $\sigma_n$ plays an important role in controlling the focus and the diversity of the particle distributions. Given the GB locations from the global maximum, we can make a number of ``good particles" at different time steps. By enforcing the likelihoods (Eq.~\ref{Eq:conditionalPDF}) of these good particles to fall within the desired range (i.e. $max_{i}w_k^i\approx{0.2}$), $\sigma_n$ is determined for every set of the data. As $\textbf{n}_k$ is the measurement noise in the observation model (Eq.~\ref{Eq:observation}), adapting $\sigma_n$ to different data sets means adapting the observation model to different data sets, which is similar to the idea of adaptive models proposed in \cite{Zhou:Adaptive}.

Since the 3D GPR data may have significant differences from one set to another, the proposed training stage can accommodate to different environmental scenarios. In this sense, the PF has more flexibility in feature description and target recognition over other approaches, such as the KF or the global maximum. After the training stage, it begins to produce outputs instead of the global maximum. By integrating the two approaches in this way, no latency results from the process of parameter initialization.

% Use this command to get the appendices' numbers in "A", "B" instead of the
% default capitalized Roman numerals ("I", "II", etc.).
% However, the capital letter form may result in awkward subsection numbers
% (such as "A-A"). Capitalized Roman numerals are the default.
%\useRomanappendicesfalse
%
%\appendices

%\section{}

\section*{Acknowledgment}
% optional entry into table of contents (if used)
%\addcontentsline{toc}{section}{Acknowledgment}
This work was sponsored by the U.S. Army's Night Vision and Electronic System's Directorate Countermine Division and Project Management Office through a grant to the U.S. Army Research Office. The authors would like to thank their collaborators at NIITEK Inc. for providing the data. The authors would also like to thank the anonymous reviewers for their valuable comments which significantly improve the presentation of this paper.

\vspace{-0.8in}

\begin{biography}
%[{\includegraphics[width=1in,height=1.25in,clip,keepaspectratio]{Li.eps}}]
{Li Tang}
received the B.S., M.S. and Ph.D. degrees in Electronic Engineering from Xidian University of China in 1997, 2000, and 2004, respectively. She is currently a postdoc at Duke University. Her research interests include image/signal processing, computer vision and pattern recognition.
\end{biography}

\vspace{-0.8in}
%\vfill

%\begin{biographynophoto}{Peter A. Torrione}
\begin{biography}{Peter A. Torrione}
received the B.S.E.E. degree from Tufts University in 1999, and the M.S.E.E. degree in Duke University in 2002. He has been working towards his Ph.D. degree at Duke University since 2004. His research interests include statistical signal processing, pattern recognition, remote sensing and image processing, etc.
\end{biography}
%\end{biographynophoto}

\vspace{-0.8in}
%\vfill

\begin{biography}{Cihat Eldeniz}
received his B.S.E.E degree from Bogazici University, Turkey, in 2006. He is now pursuing a Ph.D. at Duke University. His current research interests include statistical and digital signal processing.
\end{biography}

\vspace{-0.8in}
%\vfill

% insert where needed to balance the two columns on the last page
%\newpage

\begin{biography}{Leslie M. Collins}
received the B.S.E.E. degree from the University of Kentucky, Lexington, in 1985, and the M.S.E.E. and Ph.D. degrees in Electrical Engineering, both from the University of Michigan, Ann Arbor, in 1986 and 1995, respectively. She is an Associate Professor in the Electrical and Computer Engineering Department (ECE), Duke University. Her current research interests include statistical signal processing, subsurface sensing as well as cochlear implants.
\end{biography}


%\vfill



\begin{thebibliography}{1}

\bibitem{Gader:HMM}
P.D.~Gader, M.~Mystkowski and Yunxin Zhao, ``Landmine detection with ground penetrating radar using hidden Markov models", \emph{IEEE Transactions on Geoscience and Remote Sensing}, vol.39, no.6, pp.1231-1244, June 2001.

\bibitem{Peter:feature}
Torrione, P. and Throckmorton, C., and Collins, L., ``Performance of an Adaptive Feature-Based Processor for a Wideband Ground Penetrating Radar System", \emph{IEEE Trans. Aerospace and Electronic Systems}, vol.42, no.2, pp.644-658, April 2006.

\bibitem{Ho:Handheld}
K. C. Ho and P. D. Gader, ``A Linear Prediction Land Mine Detection Algorithm for Hand Held Ground Penetrating Radar", \emph{IEEE Transactions on Geoscience and Remote Sensing}, vol.40, no.6, pp.1374-1385, June 2002.

\bibitem{Potin:SVM}
D. Potin, Philippe Vanheeghe, E. Duflos and M. Davy, ``An abrupt change detection algorithm for buried landmines localization", \emph{IEEE transations on Geoscience and Remote Sensing}, vol.44, no.2, pp.260-272, Feb. 2006.

\bibitem{Zhu:Hyperbolas}
Quan Zhu and L. M. Collins, ``Application of feature extraction methods for landmine detection using the Wichmann/Niitek ground-penetrating radar", \emph{IEEE Transactions on Geoscience and Remote Sensing}, vol.43, no.1, pp.81-85, Jan. 2005.

\bibitem{Merwe:clutterReduction}
van der Merwe, A. and Gupta, I.J., ``A novel signal processing technique for clutter reduction in GPR measurements of small, shallow land mines", \emph{IEEE Transactions on Geoscience and Remote Sensing}, vol.38, no.6, pp.2627-2637, Nov. 2000.

\bibitem{Xu:arrayGPR}
Xiaoyin Xu, Miller, E.L. and Rappaport, C.M. and Sower, G.D., ``Statistical method to detect subsurface objects using array ground-penetrating radar data", \emph{IEEE Transactions on Geoscience and Remote Sensing}, vol.40, no.4, pp.963-976, April 2002.

\bibitem{Delbo:fuzzy}
Delbo, S. and Gamba, P. and Roccato, D., ``A fuzzy shell clustering approach to recognize hyperbolic signatures in subsurface radar images", \emph{IEEE Transactions on Geoscience and Remote Sensing}, vol.38, no.3, pp.1447-1451, May 2000.

\bibitem{Ho:frequency}
K. C. Ho, P. D. Gader and J. N. Wilson, ``Improving landmine detection using frequency domain features from ground penetrating radar", \emph{International Conference on Geoscience and Remote Sensing Symposium (IGARSS'04)}, vol.40, no.6, pp.1374-1385, Sep. 2004.

\bibitem{Carevic:Wavelet}
Carevic, D., ``Wavelet-based Method for Detection of Shallowly Buried Objects from GPR Data", \emph{Proceedings of Information, Decision and Control, (IDC99)}, pp.201-206, Feb. 1999.

\bibitem{Brunzell:shallowly}
Brunzell, H., ``Detection of shallowly buried objects using impulse radar",\emph{IEEE Transactions on Geoscience and Remote Sensing}, vol.37, no.2, pp.875-886, Mar. 1999.

\bibitem{Wu:AdaptiveGB}
Wu, R., A. Clement, J. Li, E. G. Larsson, M. Bradley, J. Habersat and G. Maksymonko, ``Adaptive Ground Bounce Removal", \emph{ELECTRONICS LETTERS}, vol.37, no.20, pp.1250-1252, 2001.

\bibitem{Rappaport:roughGround}
Rappaport, C. and El-Shenawee, M., ``Modeling GPR signal degradation from random rough ground surface", \emph{Proceedings of IEEE 2000 International Geoscience and Remote Sensing Symposium, (IGARSS 2000)}, vol.7, pp.3108-3110, July 2000.

\bibitem{Kempen:parameters}
van Kempen, L. and Sahli, H., ``Signal processing techniques for clutter parameters estimation and clutter removal in GPR data for landmine detection", \emph{Proceedings of the 11th IEEE Signal Processing Workshop on Statistical Signal Processing}, pp.158-161, Aug. 2001.

\bibitem{Gordon:Tutorial}
S. Arulampalam, S. Maskell, N. J. Gordon, and T. Clapp, ``A Tutorial on Particle Filters for On-line Non-linear/Non-Gaussian Bayesian Tracking", \emph{IEEE Transactions of Signal Processing}, vol.50, no.2, pp.174-188, Feb. 2002.

\bibitem{Welch:Kalman}
Greg Welch and Gary Bishop, ``An Introduction to the Kalman Filter", \emph{SIGGRAPH 2001}, ACM Press, Los Angeles, Aug. 2001.

\bibitem{Zhou:Adaptive}
Zhou, S.K.; Chellappa, R.; Moghaddam, B., ``Visual Tracking and Recognition Using Appearance-Adaptive Models in Particle Filters", \emph{IEEE Transactions on Image Processing}, vol.13, no.11, pp.1491-1506, Nov. 2004.

\bibitem{Andrieu:invited}
C. Andrieu, A. Doucet and V. Tadic, ``Particle Methods for Change Detection, System Identification and Control", \emph{Proceedings of the IEEE}, vol.92, no.3, Mar. 2004 (invited paper).

\bibitem{Rui:UnscentedPF}
Y. Rui and Y. Chen, ``Better proposal distributions: Object tracking using unscented particle filter", \emph{in Proc. IEEE Conf. on Computer Vision and Pattern Recognition}, Kauai, Hawaii, vol.II, pp.786-793, Dec. 2001.

\bibitem{Blake:CONDENSATION}
Michael Isard and Andrew Blake, ``CONDENSATION: conditional density propagation for visual tracking", \emph{International Journal of Computer Vision}, vol.29, no.1, pp.5-28, 1998.

\bibitem{Wang:Modular}
Ping Wang and James M. Rehg, ``A Modular Approach to the Analysis and Evaluation of Particle Filters for Figure Tracking", \emph{IEEE Conference on Computer Vision and Pattern Recognition}, pp.790-797, June 2006.

\bibitem{Gordon:Novel}
N. Gordon, D. Salmond and A. F. M. Smith, ``Novel approach to nonlinear and non-Gaussian Bayesian state estimation", \emph{Proceedings IEE-F}, vol.140, no.2, pp.107-113, April 1993.

\bibitem{Doucet:Book}
A. Doucet, N. de Freitas and N. Gordon, Sequential Monte Carlo Methods in Practice, Springer, New York, 2001.

\bibitem{Jeroen:Influence}
Jeroen Lichtenauer, Marcel Reinders, Emile Hendriks, ``Influence of The Observation Likelihood Function on Particle Filtering Performance in Tracking Applications", \emph{Sixth IEEE International Conference on Automatic Face and Gesture Recognition}, pp.767-772, May 2004.

\end{thebibliography}
\end{document}